%% file: main.tex
\documentclass{MITcsail}

\usepackage{microtype}
\usepackage{graphicx}
\usepackage{float}
\usepackage{subcaption}
\usepackage{booktabs} 
\usepackage{multirow}
\usepackage{multicol}

\usepackage{hyperref} 
\usepackage[ruled,vlined,linesnumbered]{algorithm2e}
\SetKwComment{tcc}{// }{}

\definecolor{darkred}{RGB}{140, 21, 21}
\definecolor{citegray}{gray}{0.7}
\definecolor{orange}{HTML}{F58025}
\definecolor{deepred}{rgb}{0.631,0.102,0.102}
\definecolor{amethyst}{rgb}{0.6, 0.4, 0.8}
\definecolor{darkgreen}{rgb}{0.3,0.7,0.3}
\definecolor{salmon}{RGB}{241, 150, 141}
\definecolor{mildyellow}{HTML}{FFF2CC}
\usepackage{amsmath}
\usepackage{amssymb}
\usepackage{mathtools}
\usepackage{amsthm}
\usepackage{xcolor}
\hypersetup{
    colorlinks=true,
    linkcolor=accentcolor,
    citecolor=accentcolor,
    filecolor=accentcolor,
    urlcolor=accentcolor,
}

\usepackage{tikz}
\usetikzlibrary{shapes.geometric, arrows.meta, positioning, fit, calc, decorations.pathreplacing, backgrounds}

\usepackage[capitalize,noabbrev]{cleveref}
\definecolor{light-teal}{RGB}{144, 209, 202}

\newcommand{\promptblock}[1]{%
  \begin{tcolorbox}[enhanced, breakable, colback=white, colframe=light-teal, boxrule=0.8pt,
    sharp corners, left=1em, right=1em, top=0.6em, bottom=0.6em]
    \textbf{Prompt:}~#1
  \end{tcolorbox}
}

\newcommand{\responseblock}[1]{%
  \begin{tcolorbox}[enhanced, breakable, colback=white, colframe=light-teal, boxrule=0.8pt,
    sharp corners, left=1em, right=1em, top=0.6em, bottom=0.6em]
    \textbf{Response:}~#1
  \end{tcolorbox}
}

\title{Temporally Extended Mixture-of-Experts Models}

\author{%
  {\large Zeyu Shen\textsuperscript{1}\qquad Peter Henderson\textsuperscript{1}}\\[0.45em]
  {\small \textsuperscript{1}Princeton University}%
}

\begin{document}

\include{notation}
\maketitle
\thispagestyle{firstpagestyle} 

\begin{abstract}
\textbf{\textit{Abstract.}}~Mixture-of-Experts models, now popular for scaling capacity at fixed inference speed, switch experts at nearly every token.
Once a model outgrows available GPU memory, this churn can render optimizations like offloading and pre-fetching ineffective.
We make the case that the options framework in reinforcement learning is a perfect match to tackle this problem, and argue for \emph{temporally extended mixture-of-experts} layers.
Building on the option-critic framework with deliberation costs, we add a controller to each layer that learns when to switch expert sets and which to load. By applying this to \texttt{gpt-oss-20b} with low-rank adapters and a self-distillation reward, our method reduces switch rates from over 50\% to below 5\% while retaining up to 90\% of base-model accuracy on MATH, MMLU, and MMMLU.
This shows that even existing pre-trained models can be converted to temporally extended MoEs with lightweight training, with the deliberation cost allowing model trainers to trade off switching rates against capability.
We hope this opens a principled path, grounded in the options framework, for memory-efficient serving and continual learning in ever-growing MoE models.

\vspace{0.8em}
\noindent\textbf{Correspondence}: \href{mailto:zs7353@princeton.edu}{zs7353@princeton.edu}, \href{mailto:peter.henderson@princeton.edu}{peter.henderson@princeton.edu}\\
\textbf{Links}: \href{https://princeton-polaris-lab.github.io/moe_webpage/}{Project Page} ~|~ \href{https://github.com/zeyushen-yo/rl_moe/tree/main}{Code}
\end{abstract}

\begin{figure}[H]
\centering
\includegraphics[width=\linewidth]{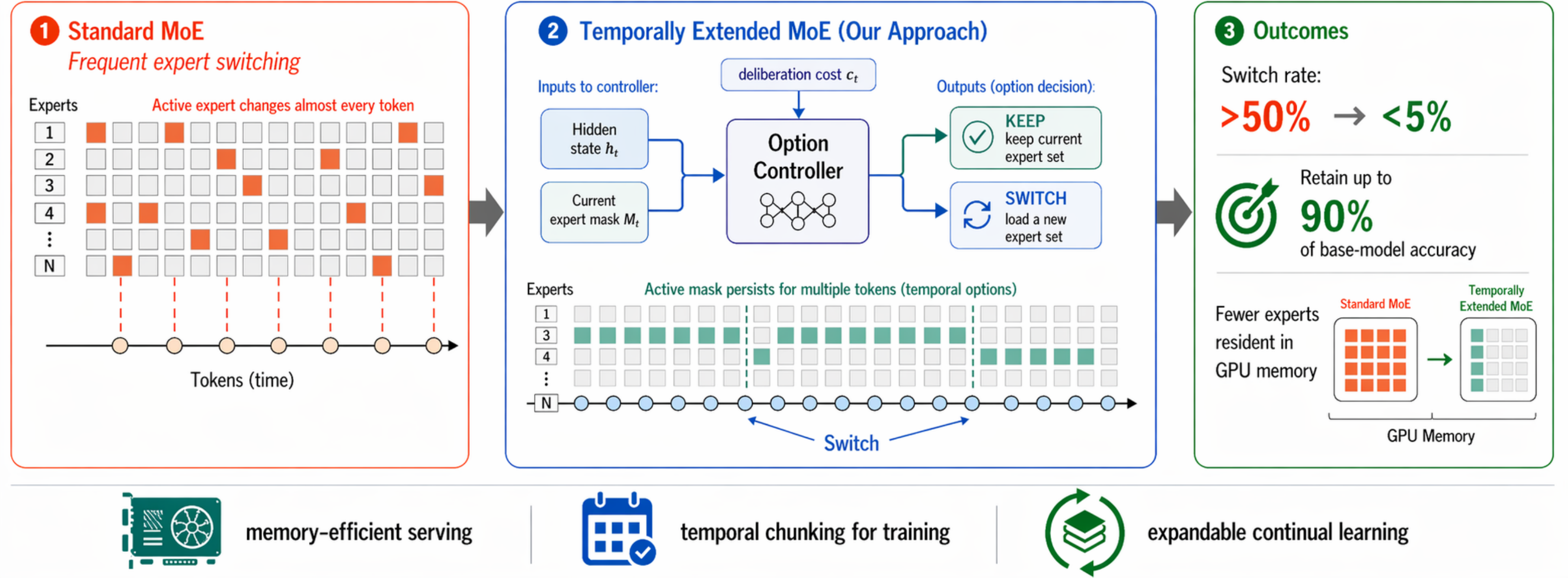}
\caption{Standard MoEs change their active expert set at almost every token. Our option controller learns when to keep the current set and when to switch, governed by a deliberation cost $\eta$. The result: switch rates collapse from over 50\% to below 5\%, while accuracy stays close to the base model — opening the door to memory-efficient serving, temporal chunking for training, and continual expansion of the expert pool.}
    \label{fig:teaser}
\end{figure}

\newpage

\input{introduction}

\input{preliminaries}

\input{motivation}

\input{method}

\input{experiments}

\input{conclusion}

\bibliography{ref}
\bibliographystyle{plainnat}

\clearpage
\beginsupplement

\input{related_works}

\input{implementation_details}

\input{additional_experiment}

\input{limitation}

\input{example}

\end{document}

%% file: notation.tex
\newcommand{\maskk}[0]{\hat{k}}
\newcommand{\activek}[0]{\tilde{k}}

%% file: introduction.tex
\section{Introduction}

Modern Large Language Models (LLMs) predominantly use some variant of Mixture-of-Experts (MoE) layers in their architecture~\citep{shazeer2017outrageouslylargeneuralnetworks, fedus2022switchtransformersscalingtrillion, Cai_2025, jiang2024mixtralexperts}, including \texttt{Gemini-2.5-Pro}~\citep{comanici2025gemini25pushingfrontier}, \texttt{GLM5}~\citep{glm5team2026glm5vibecodingagentic}, \texttt{Qwen3.5-397B-A17B}~\citep{qwen3.5}, \texttt{Qwen3-Next-80B-A3B}~\citep{qwen3next}, \texttt{DeepSeek-V3}~\citep{deepseekai2025deepseekv3technicalreport}, and \texttt{gpt-oss}~\citep{openai2025gptoss120bgptoss20bmodel}.
MoEs activate only a sparse subset of experts for each token, allowing inference-time compute to stay flat even as the total number of parameters grows. So, for example, a 120B parameter model, like \texttt{gpt-oss-120b} might only activate 5.1B parameters at a time.
In principle, with enough memory, one could add extremely large numbers of experts, while keeping inference latency the same. Scaling the number of experts could potentially come with gains in capabilities~\citep{clark2022unified,he2024mixture}. And leveraging growing numbers of experts could even be helpful for improving neuroplasticity and continual learning (though this has yet to be thoroughly explored).

However, this vision becomes challenging with finite memory resources.\footnote{At the time of writing, for example, RAM costs have \href{https://www.theverge.com/news/839353/pc-ram-shortage-pricing-spike-news}{grown} significantly due to AI-related demand, making scaling memory challenging.} Once the total number of experts outgrows GPU memory, weights must be offloaded to host memory or disk and loaded on demand~\citep{eliseev2023fastinferencemixtureofexpertslanguage, xue2025moeinfinityefficientmoeinference}. 
Each load incurs latency that would interrupt workflows and reduce throughput.
Current MoE architectures largely ignore this switching cost, assuming that all experts can be kept in memory. Across three frontier open-source MoEs, the average switch rate is large, with the active expert set changing at almost every token generation (\S\ref{sec:motivation}).

Prior work addresses memory-related challenges for MoEs mainly through two families of approaches. First, some works apply expert pruning to reduce the total number of experts by permanently removing or merging experts, possibly with additional fine-tuning~\citep{xie2024moeprunerpruningmixtureofexpertslarge, lu2024expertsequalefficientexpert, muzio2024seermoesparseexpertefficiency}. Second, some works explore caching, prefetching, and offloading-aware serving methods and design heuristics, often based on expert activations or correlations in expert usage across layers or prompts, to decide which experts to keep on GPU and which to fetch from host memory~\citep{xue2025moeinfinityefficientmoeinference, song2025promoefastmoebasedllm, tairin2025emoetaskawarememoryefficient, zhang2025duoservemoedualphaseexpertprefetch}. 

We observe, however, that this problem has a direct analogue in reinforcement learning. Choosing when to commit to a set of resources and when to pay the cost of switching is exactly the structure formalized by \emph{temporally extended actions} in the options framework~\citep{SUTTON1999181}. An agent selects a high-level ``option'' that persists over multiple time steps; switching to a new option incurs a deliberation cost~\citep{harb2017waitingoptionlearning}. 

We propose \textbf{temporally extended mixture-of-experts}, where we train a lightweight controller---a policy over options---to decide when to switch expert sets and which new set to load. The controller is optimized via the option-critic architecture~\citep{bacon2016optioncriticarchitecture} with deliberation costs~\citep{harb2017waitingoptionlearning}.
Because switching cost is an explicit term in the objective, the controller discovers temporal structure. It switches experts only when the expected quality gain justifies the cost. 

In this work we contribute the following. First, we propose the design philosophy of temporally extended MoEs as part of the options framework, identifying that excessive expert switching could lead to missed opportunities in memory optimizations across training, inference, and continual learning. 
Second, we formalize the dynamic loading of experts as a semi-Markov Decision Process (s-MDP), casting expert masks as options and expert loading latency as a deliberation cost. We adapt and apply the option-critic framework for optimization and design a lightweight controller that can modify most modern MoE architectures.
Third, we show that \texttt{gpt-oss-20b} can be trained using our option-critic method and a very small amount of adapters to reduce switching rates from over 50\% (every other token) to under 5\%, or even 1\%, with configurable performance trade-offs commensurate with the deliberation cost. This shows that temporally extended MoEs can potentially be leveraged even without large-scale pretraining.

We believe this work points toward a broader principle for MoE post-training. As expert counts continue to grow, potentially scaling with available disk rather than GPU memory, the cost of switching will increasingly dominate serving latency. Training controllers that treat expert loading as temporally extended decisions, with explicit deliberation costs, may offer a principled path for managing this trade-off. We view our framework as a first step in this direction, and present concrete evidence for its viability.

%% file: preliminaries.tex
\section{Preliminaries}
\label{sec:preliminaries}
\subsection{MDP, s-MDP, and Options}
We consider a Markov decision process (MDP) $(\mathcal{S}, \mathcal{A}, P, r, \gamma)$ with states $s \in \mathcal{S}$, actions $a \in \mathcal{A}$, transition kernel $P(s' \mid s, a)$, reward $r(s, a)$, and discount factor $\gamma \in [0, 1)$. A policy $\pi(a \mid s)$ induces a trajectory $\tau = (s_0, a_0, s_1, a_1, \ldots, s_T)$ and a return $G(\tau) = \sum_{t = 0}^{T = 1}\gamma^t r(s_t, a_t)$.

A semi-Markov decision process (s-MDP)~\citep{SUTTON1999181} generalizes an MDP by allowing actions that last a variable number of steps: at decision time $t_k$, the agent picks a high-level action, the environment evolves for a random duration $\kappa_k$, and the agent receives the cumulative reward during that duration before making the next decision.

Options are a standard framework to construct such temporally extended actions~\citep{SUTTON1999181}. An option $\omega \in \Omega$ is defined by a triple $(\mathcal{I}_\omega, \pi_\omega(a \mid s), \beta_\omega(s))$, where $\mathcal{I}_\omega \subseteq \mathcal{S}$ is an initiation set, i.e., states where the option may get started; $\pi_\omega(a \mid s)$ is an intra-option policy over primitive actions; and $\beta_\omega(s) \in [0, 1]$ is a termination function, i.e., the probability that the option terminates upon arriving in state $s$. A policy over options $\pi_\Omega(\omega \mid s)$ selects which option to start when at state $s$. We adopt the call-and-return option execution model~\citep{bacon2016optioncriticarchitecture}: An agent starts off with an initial option $\omega_0 \sim \pi_\Omega(\cdot \mid s_0)$. While option $\omega$ is active, sample primitive actions $a_t \sim \pi_\omega(\cdot \mid s_t)$. After each transition to $s_{t + 1}$, terminate the option with probability $\beta_\omega(s_{t + 1})$. If the option terminates, we sample a new option $\omega_{t + 1} \sim \pi_\Omega(\cdot \mid s_{t + 1})$; otherwise, we continue with the same option.

\subsection{MoE Routing and Expert Masks}
We focus on transformer layers where the MLP block is implemented as an MoE, comprising $N$ experts per MoE layer and $L$ total layers. For token position $t$ in layer $\ell$, the MoE router produces logits $g_t^{(\ell)} \in \mathbb{R}^N$ from which a distribution is obtained by
$$p_t^{(\ell)} = \text{softmax}(g_t^{(\ell)}).$$
In the base MoE router, a sparse top-$\activek$ set of experts is selected based on $g_t^{(\ell)}$, and the expert outputs are combined with normalized routing weights $p_t^{(\ell)}$. In our setting, routing is additionally constrained by a binary expert mask $\omega_t^{(\ell)} \in \{0, 1\}^N,$ where $\omega_{t, i}^{(\ell)} = 1$ indicates that expert $i$ is allowed at time $t$ in layer $\ell$. The top-$\activek$ expert selection is restricted to the set of allowed experts. This binary expert mask $\omega_t^{(\ell)}$ is the option in our setting. Throughout the paper, we use $\maskk$ to denote the number of allowed experts by the expert mask, and $\activek$ to denote the number of activated experts. 

We say a \textbf{switch} occurs at token position $t$ in layer $\ell$ whenever $\omega_t^{(\ell)} \neq \omega_{t-1}^{(\ell)}$, i.e., when the expert mask changes. In the options framework, this corresponds to the termination of the current option and the selection of a new one. The \textbf{switch rate} of a generated sequence of length $T$ is
$$\frac{1}{L}\sum_{\ell = 1}^L \frac{1}{T - 1}\sum_{t = 1}^{T - 1}\mathbf{1}\left[\omega_t^{(\ell)} \neq \omega_{t - 1}^{(\ell)}\right].$$
With a learned controller (Section~\ref{sec:method}), the controller selects $\omega_t^{(\ell)}$ at each position. For the base model, we define switching rates by treating the router itself as an option-selection policy: at each switch point, $\omega_t^{(\ell)}$ is set to the top-$\maskk$ experts according to $g_t^{(\ell)}$, and this mask persists until the $\activek$ activated experts at some future position are no longer fully contained in $\omega^{(\ell)}$, triggering a new switch. The base model switch rate thus measures how often the router's own selections would necessitate reloading experts, providing a reference point that our learned controller aims to reduce.

\subsection{Option-Critic Architecture}
The option-critic architecture~\citep{bacon2016optioncriticarchitecture} extends the policy gradient theorem to the options framework and allows for optimizing both the intra-option policies $\pi_\omega$ and the termination functions $\beta_\omega$ simultaneously. To derive the gradients, we first define the value of executing an action $a$ under state-option pair $(s, \omega)$ as
$$Q_U(s, \omega, a) = r(s, a) + \gamma\sum_{s'}P(s' \mid s, a)U(\omega, s'),$$
where $U(\omega, s')$ is the value of being in state $s'$ with option $\omega$ currently active:
$$U(\omega, s') = (1 - \beta_\omega(s'))Q_\Omega(s', \omega) + \beta_\omega(s')V_\Omega(s').$$
Here, $Q_\Omega(s, \omega) = \sum_a\pi_\omega(a \mid s)Q_U(s, \omega, a)$ is the value of executing option $\omega$ starting in state $s$, and $V_\Omega(s) = \sum_\omega \pi_\Omega(\omega \mid s)Q_\Omega(s, \omega)$ is the value of being in state $s$.

The option-critic framework optimizes the parameters of the intra-option policies $\pi_\omega$ (denoted by $\theta$) and the termination functions $\beta_\omega$ (denoted by $\nu$) to maximize the expected discounted return. This is realized via the following two theorems:

\begin{theorem}[Intra-Option Policy Gradient Theorem, Theorem 1 of~\citep{bacon2016optioncriticarchitecture}]
\label{thm:intra-option}
Given a set of options with stochastic intra-option policies $\pi_{\omega, \theta}$ differentiable with respect to $\theta$, the gradient of the expected discounted return with respect to $\theta$ is
$$\frac{\partial Q_\Omega(s_0, \omega_0)}{\partial \theta} = \sum_{s, \omega}\mu(s, \omega) \sum_a \frac{\partial \pi_{\omega}(a \mid s)}{\partial \theta}Q_U(s, \omega, a),$$
where $\mu(s, \omega)$ is a discounted weighting of state-option pairs along trajectories starting from $(s_0, \omega_0)$.
\end{theorem}

\begin{theorem}[Termination Gradient Theorem, Theorem 2 of~\citep{bacon2016optioncriticarchitecture}]
\label{thm:termination-gradient}
Given a set of options with stochastic termination functions $\beta_{\omega, \nu}$ differentiable with respect to $\nu$, the gradient of the expected discounted return with respect to $\nu$ is
$$\frac{\partial Q_\Omega(s_0, \omega_0)}{\partial \nu} = -\sum_{s, \omega}\mu(s, \omega)\frac{\partial \beta_{\omega}(s)}{\partial \nu}(Q_\Omega(s, \omega) - V_\Omega(s)).$$
\end{theorem}
Intuitively, $Q_\Omega(s, \omega) - V_\Omega(s)$ is the advantage of the current option relative to the expected value of switching to a new option. If the current option offers a higher value than the expected value of a newly selected option, the gradient update decreases termination probability $\beta_\omega(s)$, thereby extending the current option's duration.

%% file: motivation.tex
\section{Motivation}
\label{sec:motivation}
\subsection{Current Mixture-of-Experts LLMs are not Temporally Extended}
We first show that current MoE models are not temporally extended. We measure the switch rate of three frontier open-source MoE models --- \texttt{gpt-oss-20b} (32 experts, top-$4$), \texttt{gpt-oss-120b} (128 experts, top-$4$), and \texttt{Qwen3-Next-80B-A3B} (512 experts, top-$10$) --- on 100 prompts from each of the 10 categories in the Nemotron Post-Training Dataset v2~\citep{NemotronPostTrainingDatasetV2}. For each prompt, we generate 256 tokens with temperature $0.5$ and record which experts are activated at every token position and every layer.

\begin{figure*}[!h]
    \centering
    \begin{subfigure}[t]{0.32\linewidth}
        \centering
        \includegraphics[width=.8\linewidth]{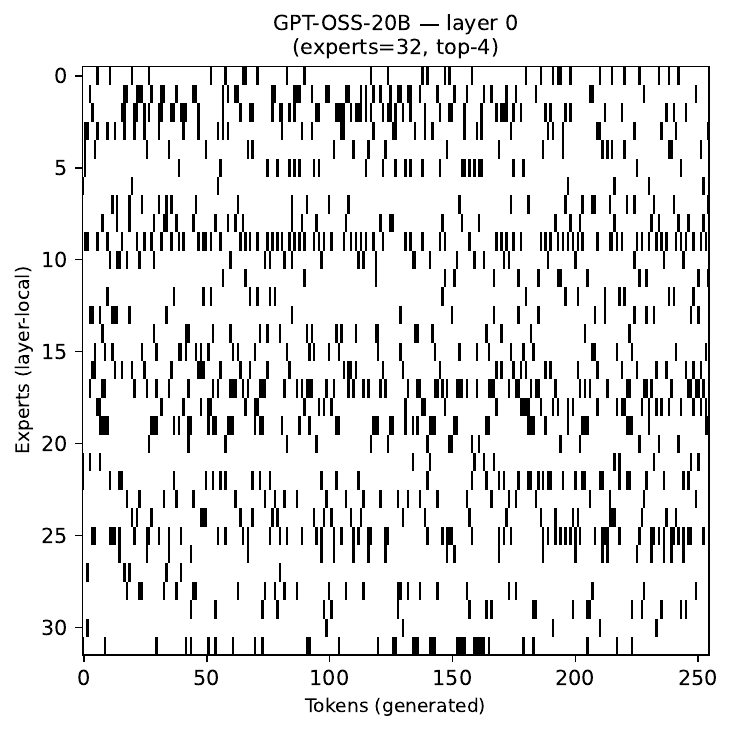}
        \caption{gpt-oss-20b}
        \label{fig:oss-20b}
    \end{subfigure}\hfill
    \begin{subfigure}[t]{0.32\linewidth}
        \centering
        \includegraphics[width=.8\linewidth]{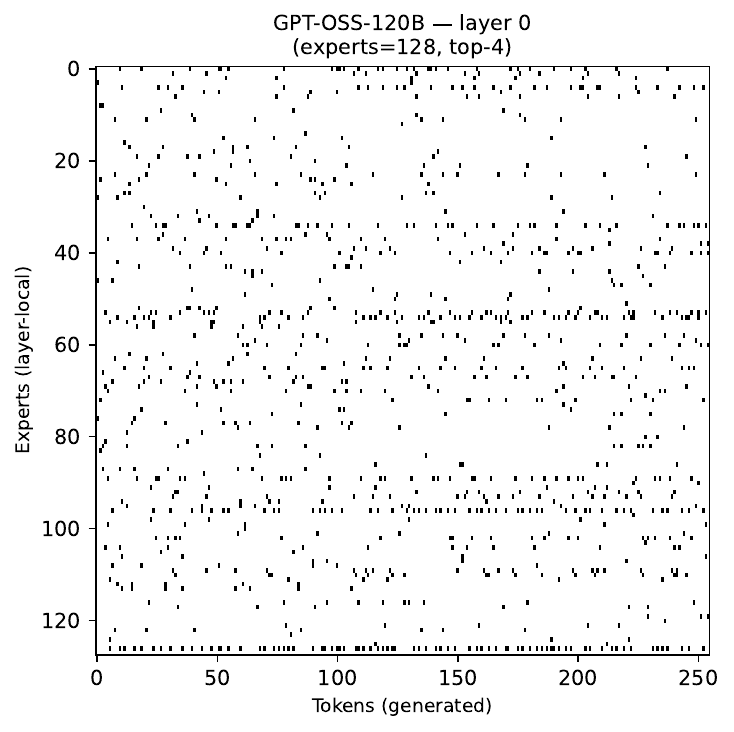}
        \caption{gpt-oss-120b}
        \label{fig:oss-120b}
    \end{subfigure}\hfill
    \begin{subfigure}[t]{0.32\linewidth}
        \centering
        \includegraphics[width=.8\linewidth]{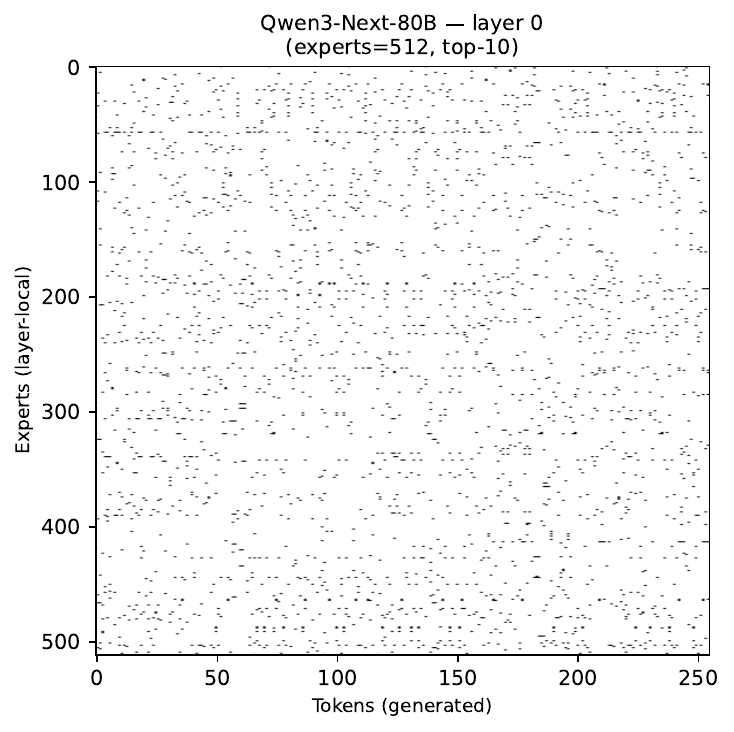}
        \caption{Qwen3-Next-80B-A3B}
        \label{fig:qwen3}
    \end{subfigure}
    \caption{Set of activated experts in layer 0 throughout the trajectory for each model}
    \label{fig:expert-profiling}
\end{figure*}

Figure~\ref{fig:expert-profiling} visualizes the expert activation pattern at layer $0$ for an arbitrarily selected prompt, where the $x$-axis is the token position and the $y$-axis are the experts. We can see  that, across all three models, expert selection shows little temporal continuity, confirmed in Table~\ref{tab:switch-rates}. The average switch rate is close to $1$ for all models, switching at nearly every token.

\begin{table*}[!h]
\centering
\resizebox{\textwidth}{!}{
\begin{tabular}{lcccccccccc}
\toprule
\textbf{Model} & \textbf{Chat} & \textbf{Code} & \textbf{Math} & \textbf{STEM} & \textbf{Multi (en)} & \textbf{Multi (de)} & \textbf{Multi (es)} & \textbf{Multi (fr)} & \textbf{Multi (it)} & \textbf{Multi (ja)} \\
\midrule
gpt-oss-20b & $0.94_{\pm0.06}$ & $0.95_{\pm0.01}$ & $0.94_{\pm0.02}$ & $0.95_{\pm0.01}$ & $0.95_{\pm0.02}$ & $0.95_{\pm0.01}$ & $0.95_{\pm0.02}$ & $0.95_{\pm0.02}$ & $0.95_{\pm0.02}$ & $0.95_{\pm0.01}$ \\
gpt-oss-120b & $0.98_{\pm0.01}$ & $0.99_{\pm0.00}$ & $0.99_{\pm0.00}$ & $0.99_{\pm0.00}$ & $0.99_{\pm0.00}$ & $0.99_{\pm0.00}$ & $0.99_{\pm0.00}$ & $0.99_{\pm0.00}$ & $0.99_{\pm0.00}$ & $0.99_{\pm0.00}$ \\
Qwen3-Next-80B & $1.00_{\pm0.00}$ & $1.00_{\pm0.00}$ & $1.00_{\pm0.00}$ & $1.00_{\pm0.00}$ & $1.00_{\pm0.00}$ & $1.00_{\pm0.00}$ & $1.00_{\pm0.00}$ & $1.00_{\pm0.00}$ & $1.00_{\pm0.00}$ & $1.00_{\pm0.00}$ \\
\bottomrule
\end{tabular}
}
\caption{Average switch rate (mean $\pm$ std) with $\maskk = \activek$ across 100 prompts per category.}
\label{tab:switch-rates}
\end{table*}

\subsection{Missed Opportunities}
The lack of temporal extension in current MoE routing leads to missed optimization opportunities across the model life-cycle. We highlight three such opportunities that temporally extended expert selection can unlock.

\paragraph{Inference serving with reduced memory.}
During autoregressive generation, standard MoE serving must keep all $N$ experts per layer resident in fast device memory (or be prepared to fetch any of them at each step), because the set of active experts is not temporally extended. When expert weights do not fit on the available GPUs, systems resort to offloading experts to host memory and prefetching them as needed~\citep{xue2025moeinfinityefficientmoeinference, song2025promoefastmoebasedllm}, but mispredictions can lead to significant latency. With temporal continuity, the active expert set is known to persist for many consecutive tokens, reducing costs from prefetching misses and enabling a simpler and more predictable serving strategy. Only the $\maskk$ active experts per layer need to reside on the GPU, and expert swaps occur only occasionally. Between switches, inference proceeds with $\frac{\maskk}{N}$ of the expert memory footprint. Since expert parameters dominate the total parameter count in modern MoE models (e.g., over 96\% in \texttt{gpt-oss-20b}), this directly translates to a substantial reduction in GPU memory requirements for serving. For example, keeping only 16 experts in $\maskk$ for \texttt{gpt-oss-20b} reduces VRAM requirements by $\sim$4.7 GiB (37\%) for 16 experts, or $\sim$7.1 GiB (55\%) for 8 experts.

\paragraph{Memory-efficient training via temporal chunking.}
A similar principle applies during training. In current MoE training pipelines, all expert parameters must be accessible during the forward and backward pass, since any token in the sequence may route to any expert. With temporally extended routing, a response can be partitioned into contiguous chunks, each associated with a fixed expert mask. Within each chunk, only the $\maskk$ experts in the current mask participate in the forward and backward computation. This opens the door to chunk-wise training strategies where inactive experts are offloaded during each chunk's forward-backward pass, reducing peak GPU memory.

\paragraph{Continual learning with expandable expert capacity.}
Temporal extension also offers a natural path toward continual learning. Because only $\maskk$ out of $N$ experts are active at any time, new experts can be added to the model without increasing the per-token compute or the active memory footprint. When adapting to a new domain or task, one can initialize fresh expert modules and let the controller learn to route to them when beneficial. The fixed active set size ($\maskk$) ensures that inference cost remains constant regardless of the new experts you added.

%% file: method.tex
\section{Method}
\label{sec:method}
In this section, we present our method towards temporally-extended control of expert routing in MoE transformers. As noted earlier, we maintain a per-layer option $\omega_t^{(\ell)}$ --- an expert mask for the allowed subset of $\maskk$ experts --- and restrict the router to select only from this set. We implement a lightweight controller that learns when to switch and which new expert mask to switch to, while simultaneously fine-tuning the MoE model's parameters via intra-option policy update.

\input{fig1}

\input{fig2}

\subsection{Options Formulation for Expert Mask Control}
For each layer $\ell$, the space of options is the combination of possible masks over experts:
\begin{equation}
\Omega^{(\ell)} = \left\{ \omega \in \{0, 1\}^N : \Vert\omega\Vert_1 = \maskk \right\}.
\end{equation}
Our controller implements call-and-return execution, an active option $\omega^{(\ell)}$ persists across tokens until a termination decision $d_t^{(\ell)} = 1$ triggers selection of a new option. During execution, the router is constrained to select only from the experts in the active option by masking all other experts' logits to $-\infty$ before the top-$\activek$ operation. In principle, one could treat the joint mask $\omega_t = (\omega_t^{(1)}, \ldots, \omega_t^{(L)})$ as a single option and learn a joint termination and selection policy over all layers. We instead factorize the controller into $L$ independent per-layer controllers for tractability, each conditioning on its layer hidden state and current mask. This yields an approximation to the joint s-MDP: each layer's controller treats the rest of the network as part of the environment while sharing the same token-level reward. Despite this approximation, it is stable to train and achieves both good performance and large switch rate reductions in practice.


\subsection{Controller Architecture}
Every MoE MLP layer $\ell$ has its own controller module, with the same architecture but separate parameters. At token $t$ and layer $\ell$, recall that $p_t^{(\ell)}$ is the softmax of the router logits, and $\omega_{t - 1}^{(\ell)}$ is the active expert mask from the previous step. Our controller operates directly on the LLM pre-MLP hidden representation $h_t^{(\ell)}$, i.e., treat $h_t^{(\ell)}$ as the state $s$. In the following, we walk through the major components of our controller.

\paragraph{Expert set embedding.} Recall that each option $\omega$ is an expert mask. To obtain a richer representation of expert masks, we use a DeepSets encoder~\citep{zaheer2018deepsets}:
\begin{equation}
z^{(\ell)}(\omega) = \frac{1}{\maskk}\sum_{i \in \omega} \varphi(e_i),
\end{equation}
where $e_i \in \mathbb{R}^{d_e}$ is a learned embedding for expert $i$ and $\varphi: \mathbb{R}^{d_e} \to \mathbb{R}^{d_c}$ is a two-layer MLP with GELU activation. Each layer has a separate encoder.

\paragraph{Termination head.} The termination decision $\beta_t^{(\ell)}$ depends on both the LLM state $h_t^{(\ell)}$ and the current option $\omega_{t - 1}^{(\ell)}$. We compute
\begin{equation}
\beta_t^{(\ell)} = \sigma\left(\mathrm{MLP}_\beta\left(\mathrm{concat}\left(\overline{h}_t^{(\ell)}, \overline{z}^{(\ell)}(\omega_{t - 1}^{(\ell)})\right)\right)\right),
\end{equation}
where $\overline{h}_t^{(\ell)} = \mathrm{RMSNorm}(h_t^{(\ell)})$ and $\overline{z}^{(\ell)} = \mathrm{RMSNorm}(z^{(\ell)})$ balance the scale of the two representations, $\mathrm{MLP}_\beta$ is a two-layer MLP with ReLU activation, and $\sigma$ is the sigmoid function. A switch decision is then sampled as $d_t^{(\ell)} \sim \mathrm{Bernoulli}(\beta_t^{(\ell)})$.

\paragraph{Value and option-value heads.}
The state-value function $V_\Omega(h_t^{(\ell)}) = w_V^\top h_t^{(\ell)} + b_V$ is a linear head on the LLM hidden states. The option-value function is
\begin{equation}
Q_\Omega(h_t^{(\ell)}, \omega) = \mathrm{MLP}_Q\left(\mathrm{concat}\left(\overline{h}_t^{(\ell)}, \overline{z}^{(\ell)}(\omega)\right)\right),
\end{equation}
where $\overline{h}_t^{(\ell)} = \mathrm{RMSNorm}(h_t^{(\ell)})$,  $\overline{z}^{(\ell)} = \mathrm{RMSNorm}(z^{(\ell)})$, and $\mathrm{MLP}_Q$ is a two-layer MLP with ReLU activation.

\paragraph{Option selection head.} When $d_t^{(\ell)} = 1$, a new option must be selected. We use a selection head $f_{\mathrm{sel}}^{(\ell)}:\mathbb{R}^d \to \mathbb{R}^N$, a linear layer initialized from the router weights, to produce candidate logits $c_t^{(\ell)} = f_{\mathrm{sel}}(h_t^{(\ell)})$. We sample $\maskk$ experts from the Plackett-Luce (PL) distribution, which defines a probability over ordered selections $(i_1, \ldots, i_{\maskk})$ by sequentially sampling without replacement:
\begin{equation}
\label{eq:pl}
P_{\mathrm{PL}}(i_1, \ldots, i_{\maskk} \mid c) = \prod_{j=1}^{\maskk} \frac{\exp(c_{i_j})}{\sum_{m \notin \{i_1, \ldots, i_{j-1}\}} \exp(c_m)}.
\end{equation} 
We denote the induced distribution as $\pi_\text{sel}(\omega \mid h)$. In real implementation, sampling sequentially is slow, so we perform sampling via the Gumbel-top-$\maskk$ trick: we add i.i.d. $\text{Gumbel}(0, 1)$ noise to $c_t^{(\ell)}$ and take the top-$\maskk$ indices of the perturbed logits, which is mathematically equivalent but fully vectorized. The new option is $\omega_t^{(\ell)} = \{i_1, \ldots, i_{\maskk}\}$. Note that we treat the sampled top-$\maskk$ indices as an ordered tuple distributed by PL; the mask is the induced unordered set, but the policy-gradient uses the ordered PL log-prob of the sampled tuple.

\paragraph{Initialization.} At $t = 0$, the initial option is set to the top-$\maskk$ experts under the router logits. The switch decision is forced to $d_0^{(\ell)} = 0$.

\subsection{Controller Training} 
We treat the non-controller parameters within the MoE model as intra-option policy. We train the controller and the MoE model using the option-critic with deliberation cost framework~\citep{harb2017waitingoptionlearning} with per-token dense rewards.

\paragraph{Reward design.} In our case, our goal is to transform a pretrained MoE into a temporally-extended MoE while maintaining its previous quality and performance. As such we follow~\citep{opd}. We use the per-token reverse KL, the divergence between the student’s and teacher’s distribution for each token conditioned on the same prior trajectory, as the per-token reward. The teacher is the original frozen MoE model (without controller and any weight updates). The student is the model we train. Specifically, 
\begin{equation}
r_t = \log p_{\mathrm{teacher}}(a_t \mid x, a_{<t}) - \log p_{\mathrm{student}}(a_t \mid x, a_{<t}),
\end{equation}
where $a_t$ is the token generated at step $t$. Note that in expectation over the student's sampling distribution, $-r_t$ is an unbiased estimator of the reverse KL divergence $\mathrm{KL}(p_{\mathrm{student}} \| p_{\mathrm{teacher}})$.

To avoid reward hacking where the trained model generates degenerate and repetitive outputs with low reverse KL, we sample tokens from a mixture of the student and teacher distributions following~\citep{gu2026minillmonpolicydistillationlarge}:
\begin{equation}
p_{\mathrm{mix}} = (1 - \tau)\, p_{\mathrm{student}} + \tau\, p_{\mathrm{teacher}},
\end{equation}
with approximate importance weights $w_t = \frac{p_{\mathrm{student}}(a_t 
\mid x, a_{<t})}{p_{\mathrm{mix}}(a_t \mid x, a_{<t})}$ applied in the policy gradient to correct for the off-policy sampling. Note that even though this is a biased approximation of the importance weights (the importance weights should have been $\prod_{t' = 1}^t\frac{p_{\mathrm{student}}(a_{t'} 
\mid x, a_{<t'})}{p_{\mathrm{mix}}(a_{t'} \mid x, a_{<t'})}$), we still use this approximation because~\citep{gu2026minillmonpolicydistillationlarge} found that it works better as it reduces the variance of gradient computation. In cases where a model is being post-trained from scratch, rather than being self-distilled into a temporally extended MoE, these rewards can be swapped for standard post-training rewards.

\paragraph{Gradient updates.} Our training procedure follows the A2OC algorithm of~\citep{harb2017waitingoptionlearning} adapted to our setting. We describe how each gradient update is instantiated.

For critic learning, we learn $V_\Omega$ and $Q_\Omega$ by minimizing squared TD errors with GAE($\lambda$)~\citep{schulman2018highdimensionalcontinuouscontrolusing} targets.  For $V_\Omega$, the TD error at token $t$ and layer $\ell$ is $\delta_t^V = r_t + \gamma V_\Omega(h_{t+1}^{(\ell)}) - V_\Omega(h_t^{(\ell)})$. For $Q_\Omega$, we bootstrap with $U(\omega, s')$ as defined in Section~\ref{sec:preliminaries}:
$\delta_t^Q = r_t + \gamma U(\omega_t^{(\ell)}, h_{t+1}^{(\ell)}) - Q_\Omega(h_t^{(\ell)}, \omega_t^{(\ell)})$,
where $\omega_t^{(\ell)}$ is the option executed at step $t$. The critics jointly minimize $\left(V_\Omega(h_t^{(\ell)}) - \hat{V}_t^{\mathrm{targ}}\right)^2 + \left(Q_\Omega(h_t^{(\ell)}, \omega_t^{(\ell)}) - \hat{Q}_t^{\mathrm{targ}}\right)^2$ respectively, with targets $\hat{V}_t^{\mathrm{targ}} = V_\Omega(h_t^{(\ell)}) + \hat{A}_t^V$ and $\hat{Q}_t^{\mathrm{targ}} = Q_\Omega(h_t^{(\ell)}, \omega_t^{(\ell)}) + \hat{A}_t^Q$ computed from the GAE advantages $\hat{A}_t^V = \sum_{j=0}^{\infty}(\gamma\lambda)^j \delta_{t+j}^V$ and $\hat{A}_t^Q = \sum_{j=0}^{\infty}(\gamma\lambda)^j \delta_{t+j}^Q$. 

For intra-option policy update, we apply Theorem~\ref{thm:intra-option} to update the intra-option policy parameters $\theta$ (expert and router parameters). In our setting, the intra-option policy $\pi_{\omega}$ is the LLM next-token-probability distribution and the primitive actions are generated tokens. Using log-derivative trick, the gradient from Theorem~\ref{thm:intra-option} can be written as
\begin{equation}
\mathbb{E}_{(s,\omega)\sim\mu,\, a\sim\pi_{\omega,\theta}} \!\left[\frac{\partial \log \pi_{\omega}(a \mid s)}{\partial \theta}\, Q_U(s, \omega, a)\right].
\end{equation}
In practice, we estimate $Q_U(s,\omega,a)$ with the Monte Carlo return $\bar{G}_t = \sum_{j\geq 0}\gamma^j r_{t+j}$.

For termination gradient update, we augment Theorem~\ref{thm:termination-gradient} with the deliberation cost $\eta$ as in~\citep{harb2017waitingoptionlearning}. The gradient of the expected return with respect to the termination parameters $\nu$ becomes
\begin{equation}
-\sum_{s,\omega} \mu(s,\omega) \frac{\partial \beta_\omega(s)}{\partial \nu} \Big(Q_\Omega(s, \omega) - V_\Omega(s) + \eta\Big),
\end{equation}
where $\mu(s,\omega)$ is the discounted state-option visitation distribution as defined in Theorem~\ref{thm:intra-option}. Here, $\eta$ serves as a margin so that termination is only preferred when the current option is sufficiently worse than alternatives to overcome the deliberation cost. 

For the option selection heads, we update them only when a switch occurs. In particular, when $d_t^{(\ell)} = 1$, we update the option selection head parameters $\phi$ using the policy gradient
\begin{equation}
\sum_{s,\omega} \mu(s,\omega) \nabla_\phi \log \pi_{\mathrm{sel}}(\omega \mid h)\, \big(Q_\Omega(s, \omega) - V_\Omega(s)\big).
\end{equation}
We present a high-level version of the algorithm in Algorithm~\ref{alg:moe_oc_simple} and the full algorithm as well as additional implementation details in Section~\ref{sec:exp-setup}.

\begin{algorithm}[!h]
\caption{MoE Option-Critic Training (High-Level)}
\label{alg:moe_oc_simple}
\KwIn{MoE model with $L$ layers, $N$ experts per layer, top-$\activek$ routing; prompt dataset $\mathcal{D}$; teacher model $p_{\mathrm{teacher}}$; teacher mixing coefficient $\tau$; discount $\gamma$; GAE parameter $\lambda$; deliberation cost $\eta$, learning rates $\alpha_{\text{controller}}$, $\alpha_\text{intra}$. We use $\theta$ to denote the LLM parameters, $\nu$ to denote the termination head, $\psi$ to denote the critic parameters (including $V_\Omega$ and $Q_\Omega$), and $\phi$ to denote the option selection head parameters.}
\For{each training iteration}{
    Sample prompt $x \sim \mathcal{D}$\;
    \tcc{Rollout with teacher mixing}
    Initialise $\omega_0^{(\ell)} \leftarrow \mathrm{TopK}(g_0^{(\ell)}, \maskk)$ from router logits, for each layer $\ell$\;
    \For{$t = 1, \ldots, T$}{
        \For{each layer $\ell$}{
            Compute termination probability $\beta_t^{(\ell)} \in [0,1]$ from hidden state $h_t^{(\ell)}$ and current option $\omega_{t-1}^{(\ell)}$\;
            Sample $d_t^{(\ell)} \sim \mathrm{Bernoulli}(\beta_t^{(\ell)})$\;
            \lIf{$d_t^{(\ell)} = 1$}{select new option $\omega_t^{(\ell)}$ by sampling $\maskk$ experts via Plackett--Luce}
            \lElse{persist $\omega_t^{(\ell)} \leftarrow \omega_{t-1}^{(\ell)}$}
            Mask router to experts in $\omega_t^{(\ell)}$\;
        }
        Sample token $a_t \sim p_{\mathrm{mix}} = (1-\tau)\,\pi_{\omega,\theta} + \tau\,p_{\mathrm{teacher}}$\;
        Record importance weight $w_t \leftarrow \pi_{\omega,\theta}(a_t) / p_{\mathrm{mix}}(a_t)$ and reward $r_t \leftarrow \log p_{\mathrm{teacher}}(a_t) - \log \pi_{\omega,\theta}(a_t)$\;
    }
    \tcc{Controller update}
    \For{each layer $\ell$}{
        Compute GAE($\lambda$) targets $\hat{V}_t^{\mathrm{targ}}$, $\hat{Q}_t^{\mathrm{targ}}$ using $V_\Omega$, $Q_\Omega$, and $r_t$\;
        \For{$t = 1, \ldots, T$}{
            Accumulate termination gradient: $d\nu \mathrel{-\!=} w_t \nabla_\nu \beta_t^{(\ell)} \bigl(Q_\Omega(h_t^{(\ell)}, \omega_{t-1}^{(\ell)}) - V_\Omega(h_t^{(\ell)}) + \eta\bigr)$\;
            Accumulate selection gradient (when $d_t^{(\ell)}\!=\!1$): $d\phi \mathrel{+\!=} w_t \nabla_\phi \log \pi_{\mathrm{sel}}(\omega_t^{(\ell)}) \bigl(Q_\Omega(h_t^{(\ell)}, \omega_t^{(\ell)}) - V_\Omega(h_t^{(\ell)})\bigr)$\;
            Accumulate critic loss: $d\psi \mathrel{-\!=} \nabla_\psi\bigl[(V_\Omega - \hat{V}_t^{\mathrm{targ}})^2 + (Q_\Omega - \hat{Q}_t^{\mathrm{targ}})^2\bigr]$\;
        }
    }
    \tcc{Intra-option policy update}
    \For{$t = 1, \ldots, T$}{
        Compute discounted return $\bar{G}_t = \sum_{j \ge 0} \gamma^j r_{t+j}$\;
        $d\theta \mathrel{+\!=} w_t \nabla_\theta \log \pi_{\omega,\theta}(a_t) \cdot \bar{G}_t$\;
    }
    $(\nu, \psi, \phi) \leftarrow (\nu, \psi, \phi) + \alpha_{\mathrm{ctrl}} \tfrac{1}{L}(d\nu, d\psi, d\phi)$;\enspace$\theta \leftarrow \theta + \alpha_{\mathrm{intra}} \, d\theta$\;
}
\end{algorithm}

%% file: fig1.tex
\definecolor{cblue}{RGB}{197,217,243}     
\definecolor{cpurple}{RGB}{212,200,238}   
\definecolor{cgreen}{RGB}{198,228,198}    
\definecolor{cgray}{RGB}{232,232,232}     
\definecolor{cborder}{RGB}{120,120,140}   

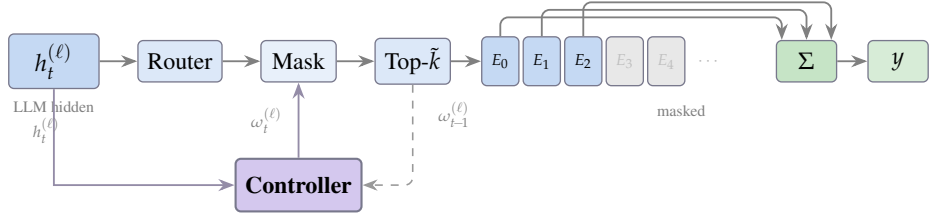
\begin{figure*}[!h]
\centering
\begin{tikzpicture}[
    >=Stealth,
    node distance=0.4cm,
    nd/.style={rectangle, draw=cborder, rounded corners=2pt, 
        minimum height=0.55cm, align=center, font=\footnotesize, line width=0.5pt},
    expert/.style={nd, minimum width=0.45cm, minimum height=0.65cm, font=\tiny},
    ctrl/.style={nd, minimum width=1.4cm, minimum height=0.7cm, 
        line width=0.8pt, font=\footnotesize\bfseries},
    arr/.style={->, black!50, line width=0.7pt},
    darr/.style={->, dashed, black!40, line width=0.6pt, rounded corners=3pt},
    lbl/.style={font=\tiny, black!50},
]

\node[nd, fill=cblue, minimum width=1.2cm] (input) {$h_t^{(\ell)}$};
\node[lbl, below=0.02cm of input] {LLM hidden};

\node[nd, fill=cblue!60, minimum width=1.1cm, right=0.5cm of input] (router) {Router};
\node[nd, fill=cblue!40, minimum width=1.0cm, right=0.5cm of router] (mask) {Mask};
\node[ctrl, fill=cpurple, below=1.0cm of mask] (controller) {Controller};
\node[nd, fill=cblue!60, minimum width=1.0cm, right=0.5cm of mask] (topk) {Top-$\activek$};

\node[expert, fill=cblue, right=0.4cm of topk] (e0) {$E_0$};
\node[expert, fill=cblue, right=0.04cm of e0] (e1) {$E_1$};
\node[expert, fill=cblue, right=0.04cm of e1] (e2) {$E_2$};
\node[expert, fill=cgray, right=0.04cm of e2] (e3) {\textcolor{black!25}{$E_3$}};
\node[expert, fill=cgray, right=0.04cm of e3] (e4) {\textcolor{black!25}{$E_4$}};
\node[right=0.03cm of e4, font=\tiny, black!30] {$\cdots$};

\node[nd, fill=cgreen, minimum width=0.8cm, right=1.2cm of e4] (sum) {$\Sigma$};
\node[nd, fill=cgreen!70, minimum width=0.8cm, right=0.4cm of sum] (output) {$y$};

\draw[arr] (input) -- (router);
\draw[arr] (router) -- (mask);
\draw[arr] (mask) -- (topk);
\draw[arr] (topk) -- (e0.west);
\draw[arr, rounded corners=3pt] (e0.north) -- ++(0, 0.25) -| ($(sum.north west)+(0.08,0)$);
\draw[arr, rounded corners=3pt] (e1.north) -- ++(0, 0.35) -| (sum.north);
\draw[arr, rounded corners=3pt] (e2.north) -- ++(0, 0.45) -| ($(sum.north east)+(-0.08,0)$);
\draw[arr] (sum) -- (output);

\draw[arr, cpurple!70!black] (input.south) |- (controller.west);
\node[lbl] at ($(input.south)+(-0.1, -0.55)$) {$h_t^{(\ell)}$};

\draw[arr, cpurple!70!black] (controller.north) -- (mask.south);
\node[lbl, left=0.08cm] at ($(controller.north)!0.5!(mask.south)$) {$\omega_t^{(\ell)}$};

\draw[darr] (topk.south) |- (controller.east);
\node[lbl] at ($(topk.south)+(0.55, -0.4)$) {$\omega_{t\!-\!1}^{(\ell)}$};

\node[lbl, below=0.12cm of e4, xshift=0.2cm] {masked};

\end{tikzpicture}
\caption{Controller integration with MoE layer. The controller observes LLM hidden states $h_t^{(\ell)}$ and the current option $\omega_{t-1}^{(\ell)}$, and outputs a new expert mask $\omega_t^{(\ell)}$ that restricts which experts can be selected by top-$\activek$ routing. Grayed experts are masked out.}
\label{fig:controller-moe}
\end{figure*}

%% file: fig2.tex
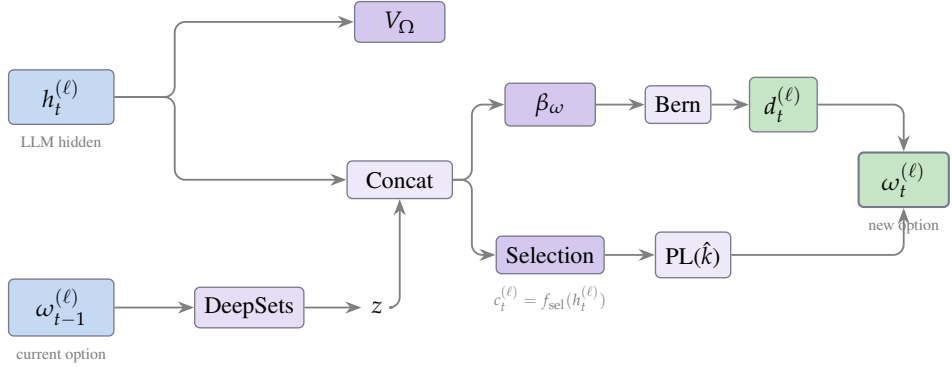
\begin{figure*}[!h]
\centering
\begin{tikzpicture}[
    >=Stealth,
    nd/.style={rectangle, draw=cborder, rounded corners=2pt, 
        minimum height=0.5cm, align=center, font=\footnotesize, line width=0.5pt},
    arr/.style={->, black!50, line width=0.7pt},
    lbl/.style={font=\tiny, black!50},
]

\node[nd, fill=cblue, minimum width=1.4cm] (h_input) at (0, 0) {$h_t^{(\ell)}$};
\node[lbl, below=0.02cm of h_input] {LLM hidden};

\node[nd, fill=cblue, minimum width=1.4cm] (omega_input) at (0, -2.8) {$\omega_{t-1}^{(\ell)}$};
\node[lbl, below=0.02cm of omega_input] {current option};

\node[nd, fill=cpurple!60, minimum width=1.4cm] (deepsets) at (2.5, -2.8) {DeepSets};

\node[font=\footnotesize] (s_label) at (4.2, -2.8) {$z$};

\node[nd, fill=cpurple!40, minimum width=1.4cm] (concat) at (4.5, -1.1) {Concat};

\node[nd, fill=cpurple, minimum width=1.2cm] (vhead) at (4.5, 1.0) {$V_\Omega$};

\node[nd, fill=cpurple, minimum width=1.2cm] (termhead) at (6.5, -0.1) {$\beta_\omega$};
\node[nd, fill=cpurple!40, minimum width=0.8cm] (bern) at (8.2, -0.1) {Bern};
\node[nd, fill=cgreen, minimum width=0.9cm] (d_out) at (9.6, -0.1) {$d_t^{(\ell)}$};

\node[nd, fill=cpurple, minimum width=1.2cm] (selhead) at (6.5, -2.1) {Selection};
\node[lbl, below=0.02cm of selhead] {$c_t^{(\ell)} = f_{\mathrm{sel}}(h_t^{(\ell)})$};
\node[nd, fill=cpurple!40, minimum width=1.0cm] (plsample) at (8.4, -2.1) {PL($\maskk$)};

\node[nd, fill=cgreen, minimum width=1.2cm, line width=0.8pt] (omega_out) at (11.2, -1.1) {$\omega_t^{(\ell)}$};
\node[lbl, below=0.02cm of omega_out] {new option};

\draw[arr] (omega_input) -- (deepsets);
\draw[arr] (deepsets) -- (s_label);

\draw[arr, rounded corners=4pt] (h_input.east) -- ++(0.8, 0) |- (vhead.west);
\draw[arr, rounded corners=4pt] (h_input.east) -- ++(0.8, 0) |- (concat.west);
\draw[arr, rounded corners=4pt] (s_label.east) -| (concat.south);

\draw[arr, rounded corners=4pt] (concat.east) -- ++(0.25, 0) |- (termhead.west);
\draw[arr, rounded corners=4pt] (concat.east) -- ++(0.25, 0) |- (selhead.west);

\draw[arr] (termhead) -- (bern);
\draw[arr] (bern) -- (d_out);

\draw[arr] (selhead) -- (plsample);

\draw[arr, rounded corners=4pt] (d_out.east) -| (omega_out.north);
\draw[arr, rounded corners=4pt] (plsample.east) -| (omega_out.south);

\end{tikzpicture}
\caption{Controller architecture. The controller takes LLM hidden states $h_t^{(\ell)}$ and the current option $\omega_{t-1}^{(\ell)}$ (a set of $\maskk$ expert indices) as inputs. A DeepSets encoder maps $\omega_{t-1}^{(\ell)}$ to a permutation-invariant set embedding $z$, which is concatenated with $h_t^{(\ell)}$ (after RMSNorm) and fed to the termination head $\beta_\omega$ and the selection head. The state-value head $V_\Omega$ uses $h_t^{(\ell)}$ alone. The selection head is a linear layer initialized from the router weights and produces candidate logits $c_t^{(\ell)} = f_{\mathrm{sel}}(h_t^{(\ell)})$, from which $\maskk$ experts are sampled via the Plackett--Luce distribution. If the Bernoulli sample $d_t^{(\ell)}\!=\!1$, the newly sampled option replaces the current one; otherwise the current option persists.}
\label{fig:controller-internal}
\end{figure*}

%% file: experiments.tex
\section{Experiments}
\label{sec:exp}
In this section, we present our main experimental setup and results.

\paragraph{Training details.} We conduct all experiments on \texttt{gpt-oss-20b}~\citep{openai2025gptoss120bgptoss20bmodel}, a Mixture-of-Experts LLM with 24 transformer layers, 32 experts per layer, and top-$4$ routing ($\activek = 4$). The model uses MXFP4 quantization natively and is dequantized to bf16 for training. We train on $4 \times$ NVIDIA 140GB H200 GPUs, using a modified version of the TRL library~\citep{vonwerra2020trl}.

For training, we use the following hyperparameters: discount factor $\gamma = 0.95$, GAE parameter $\lambda = 0.95$, value loss coefficient $0.01$. The controller is trained with learning rates $\alpha_{\text{controller}} = 10^{-4}$ using AdamW. For the intra-option policy update, we apply LoRA~\citep{hu2021loralowrankadaptationlarge} with rank $r = 16$ and $\alpha = 16$ to both expert parameters and attention parameters. Router weights are also trainable. The intra-option policy learning rate is $\alpha_{\text{intra}} = 2 \times 10^{-4}$. Each training step uses 16 prompts with a maximum prompt length of 512 tokens and max response length of 512 tokens. Token generation uses temperature $1.0$ and $\text{Top-}p = 0.95$. The teacher mixing ratio is $\tau = 0.2$.

\paragraph{Datasets and benchmarks.} For training, we use the Nemotron Post-Training Dataset v2~\citep{NemotronPostTrainingDatasetV2}, which contains prompts across 10 categories: chat, code, math, STEM, and multilingual (English, German, Spanish, French, Italian, Japanese). We use all categories. For evaluation, we evaluate on 200 randomly selected questions from MATH dataset~\citep{hendrycks2021measuringmathematicalproblemsolving}, MMLU, and MMMLU~\citep{hendrycks2021measuringmassivemultitasklanguage}, respectively. For MATH, we check correctness using the \texttt{is\_equiv()} function provided in the official github repository of~\citep{hendrycks2021measuringmathematicalproblemsolving}. All evaluations use temperature $0.5$, $\text{Top-}p = 0.95$, and max response length of $2048$ tokens. Random seed is $42$ across all experiments.

\paragraph{Baselines.} We compare our trained controller with four pruning baselines: frequency-based selection, reconstruction loss minimization~\citep{lu2024expertsequalefficientexpert}, random selection, and Wanda (structured)~\citep{sun2024simpleeffectivepruningapproach}. For all baselines, we use $128$ prompts randomly drawn from the Nemotron Post-training Dataset v2 as the calibration set---\citet{lu2024expertsequalefficientexpert} found 128 sequences to be best for calibration.\footnote{~\citet{lu2024expertsequalefficientexpert} use 128 text pieces drawn from C4 pretraining corpus (without any model generated responses) as the calibration set, which we found to have very poor performance in our setting, so we switch to Nemotron Post-training Dataset v2 and include model generated responses.} For each prompt, we generate a response with \texttt{gpt-oss-20b} so that the prompt length plus the response length does not exceed $2048$. All other configurations follow~\citep{lu2024expertsequalefficientexpert}. For frequency-based selection, we keep the $\maskk$ experts that are most frequently used on the calibration set. For reconstruction loss minimization, we keep a set of $\maskk$ experts that minimizes a reconstruction loss. We provide additional details about our implementation of reconstruction loss minimization in Section~\ref{sec:exp-setup}. For random selection, we randomly select a set of $\maskk$ experts at each token. For Wanda (structured), we conduct structured weight pruning and prune out $\frac{N - \maskk}{N}$ of the weights following~\citep{sun2024simpleeffectivepruningapproach}.\footnote{We do not compare with caching and prefetching approaches such as MoE-infinity~\citep{xue2025moeinfinityefficientmoeinference}, because they optimize the memory management of expert weights during inference rather than alter the model's routing decisions. Every expert selected by the router is still computed, with cache misses only incurring additional latency. This is different from our setting where we want to study utility-efficiency tradeoff. In fact, \citet{xue2025moeinfinityefficientmoeinference}'s work can directly be used with our method to allow for off-loading and on-loading of experts when switching options.}


\begin{table*}[!h]
\centering
\resizebox{\textwidth}{!}{
\begin{tabular}{lccccccccc}
\toprule
& & \multicolumn{4}{c}{\textbf{Pruning Baselines}} & & \multicolumn{3}{c}{\textbf{Ours (Learned Controller)}} \\
\cmidrule(lr){3-6} \cmidrule(lr){8-10}
\multirow{-2}{*}{\textbf{Benchmark}} & \multirow{-2}{*}{\textbf{Base Model}} & \textbf{Frequency} & \textbf{Reconstruction} & \textbf{Random} & \textbf{Wanda} & & \textbf{$\eta{=}0.02$} & \textbf{$\eta{=}0.03$} & \textbf{$\eta{=}0.04$} \\
\midrule
MATH  & $71.5_{\pm5.9}$ & $53.5_{\pm6.9}$ & $51.5_{\pm6.9}$ & $15.0_{\pm4.9}$ & $3.5_{\pm2.5}$  & & $\mathbf{64.0}_{\pm6.7}$ & $58.5_{\pm6.9}$ & $55.0_{\pm6.9}$ \\
{\scriptsize\color{gray} switch \%} & {\scriptsize\color{gray} $58.6_{\pm0.51}$} & & & & & & {\scriptsize\color{gray} $4.1_{\pm0.02}$} & {\scriptsize\color{gray} $1.3_{\pm0.02}$} & {\scriptsize\color{gray} $1.2_{\pm0.02}$} \\
\midrule
MMLU  & $79.5_{\pm5.7}$ & $55.5_{\pm6.9}$ & $35.0_{\pm6.7}$ & $33.5_{\pm6.5}$ & $9.0_{\pm3.9}$  & & $\mathbf{72.5}_{\pm6.3}$ & $67.5_{\pm6.5}$ & $63.0_{\pm6.7}$ \\
{\scriptsize\color{gray} switch \%} & {\scriptsize\color{gray} $57.1_{\pm0.53}$} & & & & & & {\scriptsize\color{gray} $4.2_{\pm0.02}$} & {\scriptsize\color{gray} $1.3_{\pm0.02}$} & {\scriptsize\color{gray} $1.2_{\pm0.02}$} \\
\midrule
MMMLU & $67.5_{\pm6.5}$ & $42.0_{\pm6.9}$ & $48.0_{\pm6.9}$ & $24.0_{\pm5.9}$ & $7.0_{\pm3.5}$  & & $\mathbf{59.5}_{\pm6.9}$ & $56.5_{\pm6.9}$ & $49.5_{\pm6.9}$ \\
{\scriptsize\color{gray} switch \%} & {\scriptsize\color{gray} $54.5_{\pm0.51}$} & & & & & & {\scriptsize\color{gray} $4.2_{\pm0.02}$} & {\scriptsize\color{gray} $1.4_{\pm0.02}$} & {\scriptsize\color{gray} $1.2_{\pm0.02}$} \\
\bottomrule
\end{tabular}
}
\caption{Accuracy (\%, mean $\pm$ 95\% CI) and switch rate (\%, mean $\pm$ 95\% CI) with $\maskk = 16$. ``Ours'' denotes our trained controller with deliberation cost $\eta$.}
\label{tab:results-16experts}
\end{table*}

\begin{table*}[!h]
\centering
\resizebox{\textwidth}{!}{
\begin{tabular}{lccccccccc}
\toprule
& & \multicolumn{4}{c}{\textbf{Pruning Baselines}} & & \multicolumn{3}{c}{\textbf{Ours (Learned Controller)}} \\
\cmidrule(lr){3-6} \cmidrule(lr){8-10}
\multirow{-2}{*}{\textbf{Benchmark}} & \multirow{-2}{*}{\textbf{Base Model}} & \textbf{Frequency} & \textbf{Reconstruction} & \textbf{Random} & \textbf{Wanda} & & \textbf{$\eta{=}0.02$} & \textbf{$\eta{=}0.03$} & \textbf{$\eta{=}0.04$} \\
\midrule
MATH  & $71.5_{\pm5.9}$ & $11.5_{\pm4.3}$ & $7.5_{\pm3.5}$  & $0.0_{\pm0.0}$  & $0.0_{\pm0.0}$ & & $\mathbf{27.5}_{\pm6.1}$ & $23.0_{\pm5.9}$ & $15.5_{\pm4.9}$ \\
{\scriptsize\color{gray} switch \%} & {\scriptsize\color{gray} $79.0_{\pm0.39}$} & & & & & & {\scriptsize\color{gray} $9.2_{\pm0.14}$} & {\scriptsize\color{gray} $7.4_{\pm0.12}$} & {\scriptsize\color{gray} $5.4_{\pm0.10}$} \\
\midrule
MMLU  & $79.5_{\pm5.7}$    & $12.5_{\pm4.5}$ & $2.5_{\pm2.2}$  & $4.0_{\pm2.7}$  & $0.0_{\pm0.0}$ & & $\mathbf{48.5}_{\pm6.9}$ & $41.0_{\pm6.9}$ & $38.0_{\pm6.7}$ \\
{\scriptsize\color{gray} switch \%} & {\scriptsize\color{gray} $77.4_{\pm0.45}$} & & & & & & {\scriptsize\color{gray} $8.5_{\pm0.10}$} & {\scriptsize\color{gray} $7.6_{\pm0.08}$} & {\scriptsize\color{gray} $5.0_{\pm0.06}$} \\
\midrule
MMMLU & $67.5_{\pm6.5}$    & $8.5_{\pm3.9}$  & $1.0_{\pm1.4}$  & $3.0_{\pm2.4}$  & $0.0_{\pm0.0}$ & & $\mathbf{39.0}_{\pm6.5}$ & $31.5_{\pm6.3}$ & $22.5_{\pm5.9}$ \\
{\scriptsize\color{gray} switch \%} & {\scriptsize\color{gray} $75.5_{\pm0.43}$} & & & & & & {\scriptsize\color{gray} $9.0_{\pm0.14}$} & {\scriptsize\color{gray} $8.0_{\pm0.10}$} & {\scriptsize\color{gray} $5.4_{\pm0.08}$} \\
\bottomrule
\end{tabular}
}
\caption{Accuracy (\%, mean $\pm$ 95\% CI) and switch rate (\%, mean $\pm$ 95\% CI) with $\maskk = 8$. ``Ours'' denotes our trained controller with deliberation cost $\eta$.}
\label{tab:results-8experts}
\end{table*}

\begin{figure*}[!h]
    \centering
    \begin{subfigure}[t]{0.49\linewidth}
        \centering
        \includegraphics[width=\linewidth]{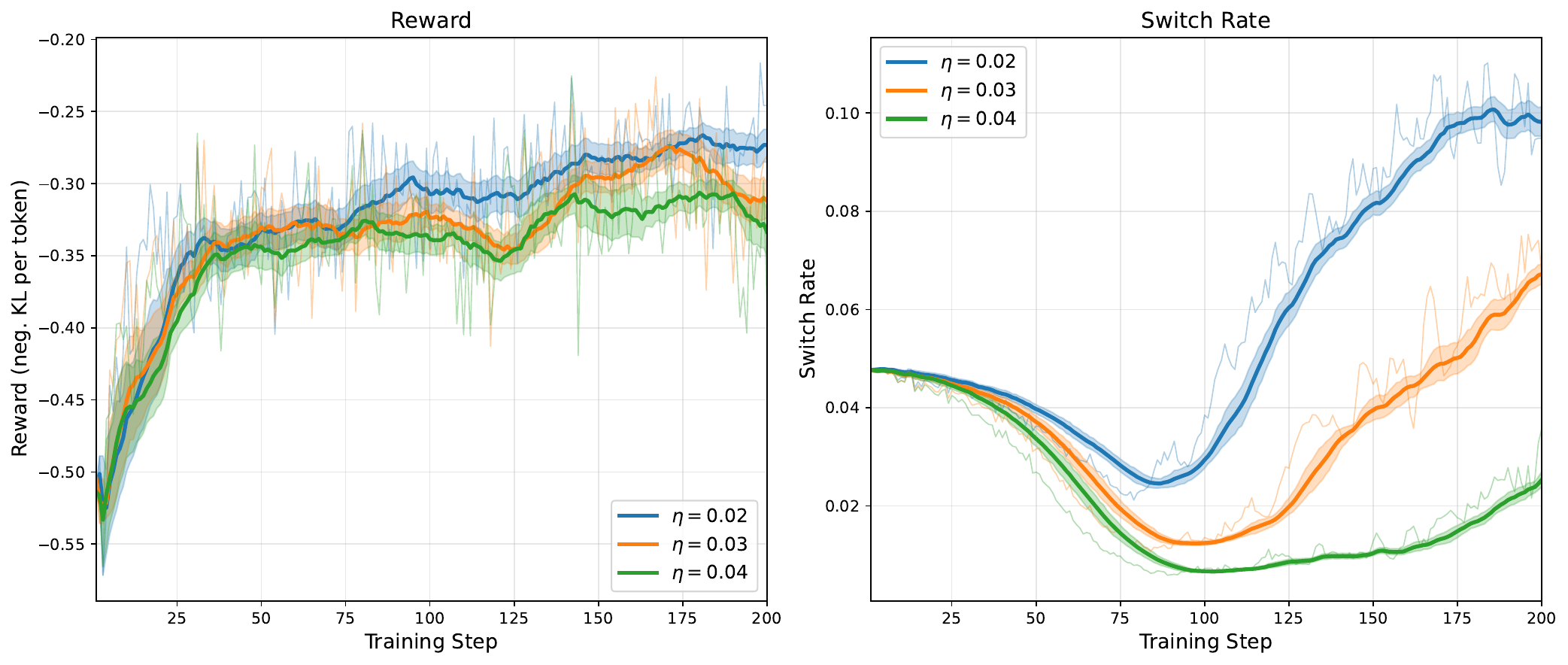}
        \caption{Training curves for $\maskk = 8$.}
        \label{fig:reward_exp8}
    \end{subfigure}\hfill
    \begin{subfigure}[t]{0.49\linewidth}
        \centering
        \includegraphics[width=\linewidth]{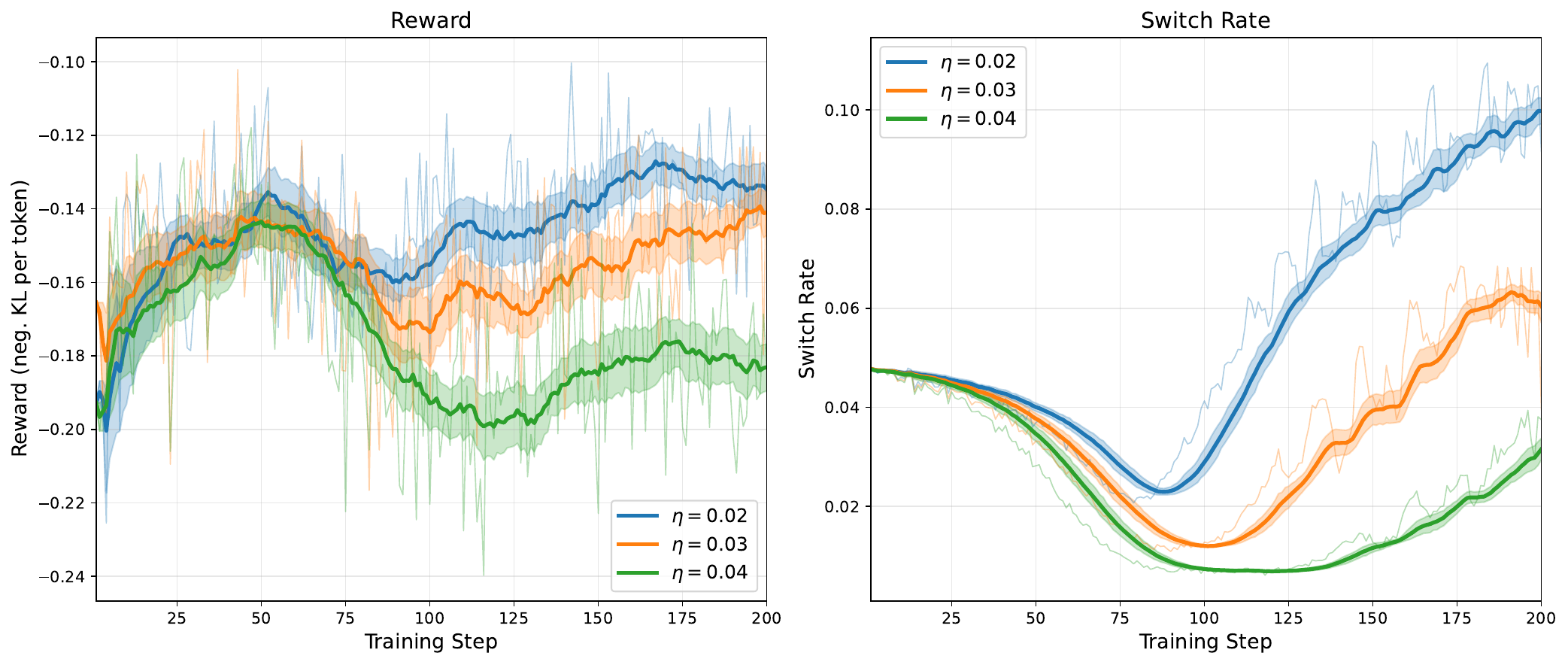}
        \caption{Training curves for $\maskk = 16$}
        \label{fig:reward_exp16}
    \end{subfigure}
    \caption{Reward curves and switch rate curves for one training run. We present the running averages over sliding windows of size $20$. The shaded bands show bootstrap 95\% confidence intervals for the running mean, computed by resampling (with replacement, 1000 times) within a sliding window of 20 steps and taking the 2.5th and 97.5th percentiles of the resampled means.}
    \label{fig:training_curve}
\end{figure*}

\paragraph{Training Dynamics.} We train the controller with varying deliberation costs $\eta \in \{0.02, 0.03, 0.04\}$ and expert budgets $\maskk \in \{8, 16\}$. The training curves of our runs are presented in Figure~\ref{fig:training_curve}.\footnote{Since we report a single training run per configuration, the observed variance in the curves reflects within-run stochasticity rather than across-run stability; we do not claim one setting trains more stably than another, only that we are able to yield a well-performing temporally extended MoE in at least one run.} Across different configurations, the reward steadily increases during training, with more pronounced gains under $\maskk = 8$. The switch rate initially decreases (as the value networks $V_\Omega, Q_\Omega$ are learning) and gradually stabilizes at a level determined by $\eta$, with higher deliberation costs yielding lower converged switch rates. Perplexity also decreases throughout training, with clearer improvements for $\maskk = 8$.

\paragraph{Benchmark Evaluation.} For $\maskk = 8$, we evaluate the checkpoint at step $300$. For $\maskk = 16$, we evaluate the checkpoint at step $120$. With $\maskk = 16$ and $\eta = 0.02$, our controller achieves accuracy close to the unpruned base model and substantially outperforms all baselines across benchmarks. Performance shows a trade-off in performance commensurate with the deliberation cost and the size of the mask $\maskk$. This trade-off might improve with a full post-training run and can be calibrated with the deliberation cost.

\begin{figure*}[!h]
    \centering
    \begin{subfigure}[t]{0.32\linewidth}
        \centering
        \includegraphics[width=.8\linewidth]{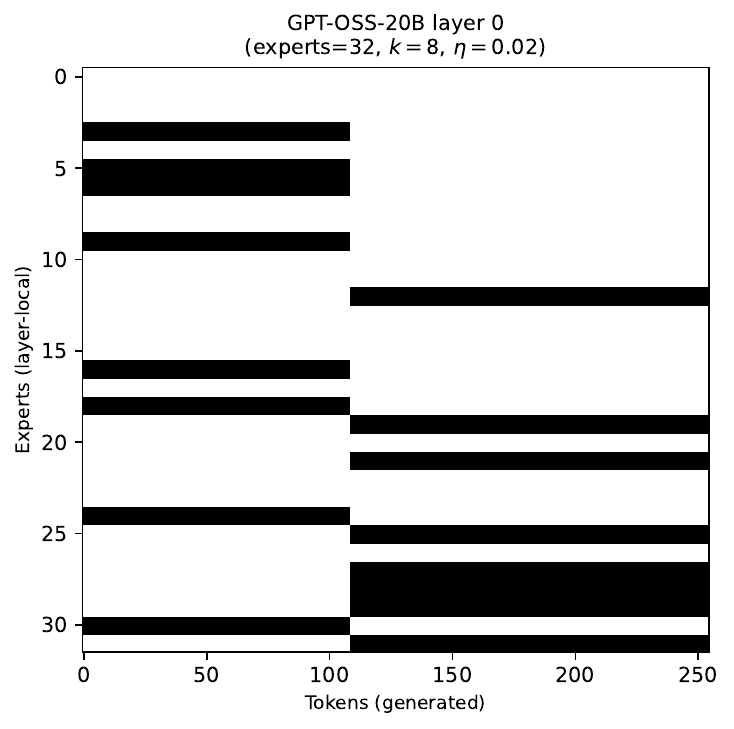}
        \caption{Layer 0}
        \label{fig:k8eta0.02layer0}
    \end{subfigure}\hfill
    \begin{subfigure}[t]{0.32\linewidth}
        \centering
        \includegraphics[width=.8\linewidth]{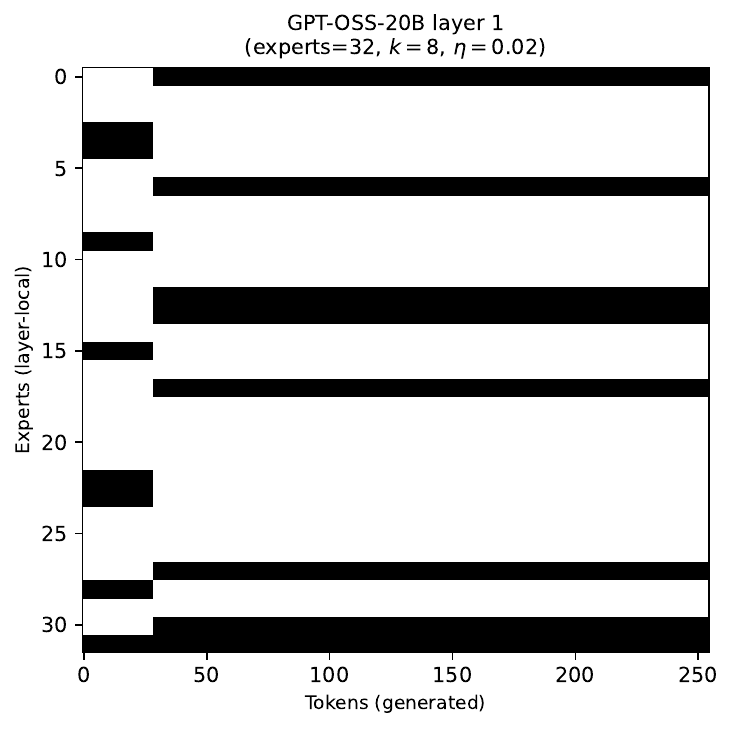}
        \caption{Layer 1}
        \label{fig:k8eta0.02layer1}
    \end{subfigure}\hfill
    \begin{subfigure}[t]{0.32\linewidth}
        \centering
        \includegraphics[width=.8\linewidth]{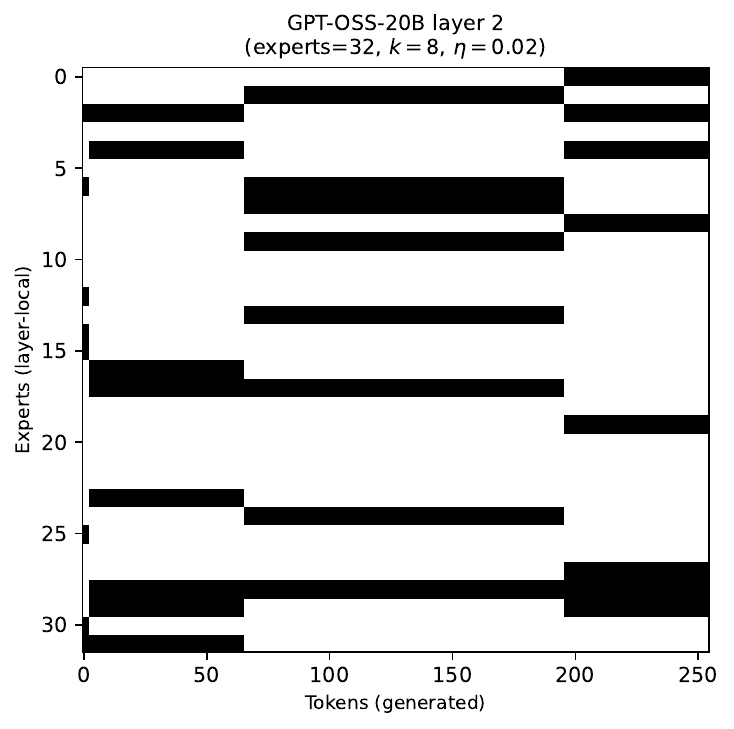}
        \caption{Layer 2}
        \label{fig:k8eta0.02layer2}
    \end{subfigure}
    \caption{Options in layer 0, 1, 2 throughout the trajectory with gpt-oss-20b, $\maskk = 8, \eta = 0.02$.}
    \label{fig:temporal_continuity_k8_eta0.02}
\end{figure*}

\begin{figure*}[!h]
    \centering
    \begin{subfigure}[t]{0.32\linewidth}
        \centering
        \includegraphics[width=.8\linewidth]{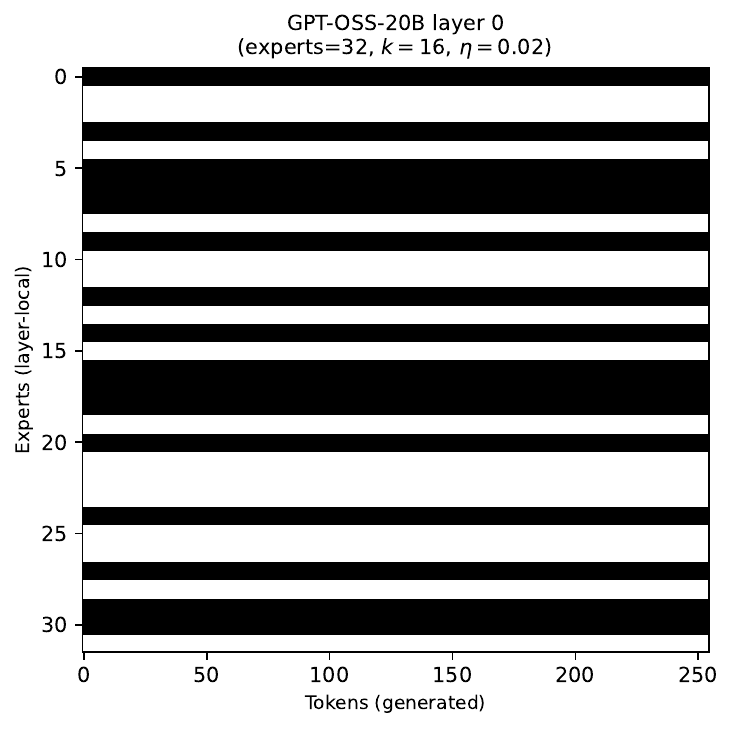}
        \caption{Layer 0}
        \label{fig:k16eta0.02layer0}
    \end{subfigure}\hfill
    \begin{subfigure}[t]{0.32\linewidth}
        \centering
        \includegraphics[width=.8\linewidth]{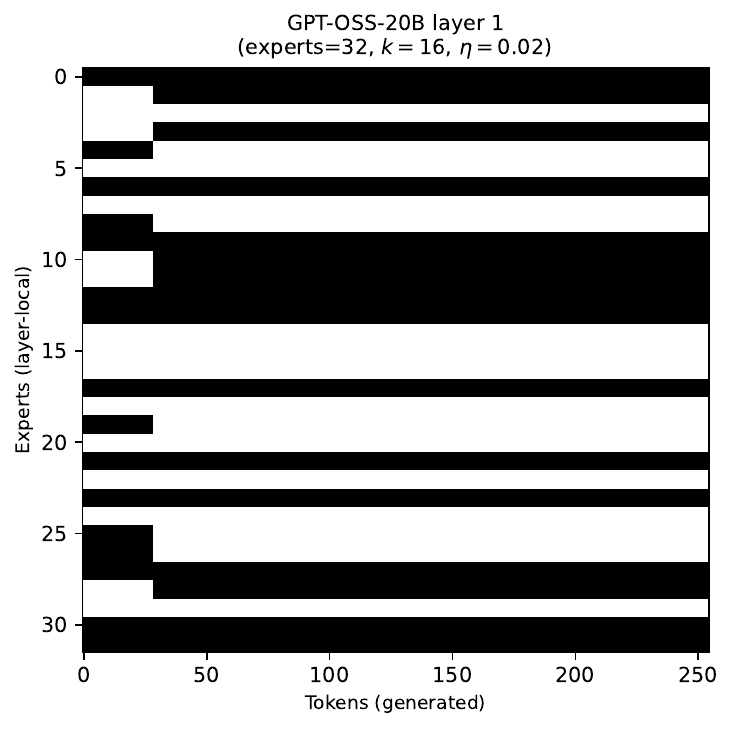}
        \caption{Layer 1}
        \label{fig:k16eta0.02layer1}
    \end{subfigure}\hfill
    \begin{subfigure}[t]{0.32\linewidth}
        \centering
        \includegraphics[width=.8\linewidth]{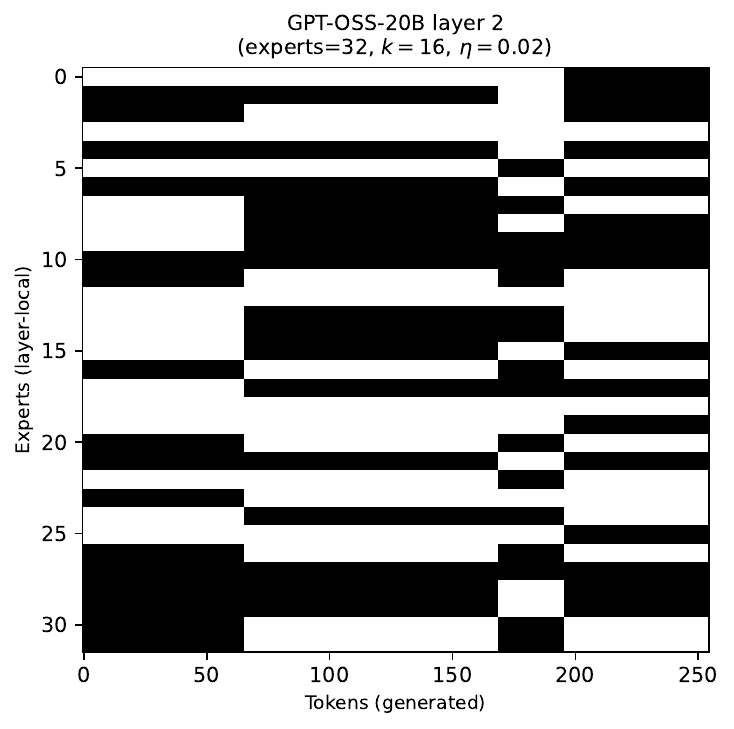}
        \caption{Layer 2}
        \label{fig:k16eta0.02layer2}
    \end{subfigure}
    \caption{Options in layer 0, 1, 2 throughout the trajectory with gpt-oss-20b, $\maskk = 16, \eta = 0.02$.}
    \label{fig:temporal_continuity_k16_eta0.02}
\end{figure*}

\paragraph{Temporal continuity with controller.} As a direct comparison with Figure~\ref{fig:expert-profiling}, we plot the option (i.e., expert mask) of \texttt{gpt-oss-20b} under our trained controller (with $\eta = 0.02$) for the same prompt used in Section~\ref{sec:motivation} throughout the generated trajectory in layer 0, 1, 2, respectively, in Figures~\ref{fig:temporal_continuity_k8_eta0.02} and~\ref{fig:temporal_continuity_k16_eta0.02}, where $x$-axis is the token position and $y$-axis are the experts. We still generate 256 tokens with temperature $0.5$ and record the expert mask at every token position and every layer. We can see that the expert selection shows significantly stronger temporal continuity. Note that different layers can have different levels of temporal continuity. We present additional plots under controller trained with $\eta = 0.03, 0.04$ in Section~\ref{sec:additional_experiment}, which show similar patterns.

We present additional analysis of training and results, including loss curves, additional switch rate analysis, repetition and perplexity analysis, in Section~\ref{sec:additional_experiment}, and concrete examples of model responses with different methods in Section~\ref{sec:examples}.

%% file: conclusion.tex
\section{Discussion and Conclusion}
In this work, we introduced the concept of temporally extended MoE models and presented a framework that addresses dynamic expert loading using the options framework.\footnote{We provide additional discussions of limitations and future directions in Section~\ref{sec:limitations}.} Our method effectively balances generation quality with the latency cost of expert set transfer. Our findings also point to a promising future direction of designing MoE architectures in a temporally extended way by making it a core objective during post-training, and potentially even during pre-training. Developing such inherently temporally extended MoE models could minimize expert switches by design, further closing the gap between massive model capacity and low-memory, low-latency serving. 

%% file: related_works.tex
\section{Related Works}
\label{app:related_works}
\paragraph{Mixture-of-Experts Models.} MoE architectures have become the dominant paradigm for scaling the capacity for LLMs. Recent advancements in MoE architectures have shifted towards a high-sparsity regime, where the total number of experts far exceeds the number of active experts per layer. For instance,~\citep{openai2025gptoss120bgptoss20bmodel} has 128 experts per layer but only activates 4 per token;~\citep{qwen3next} has 512 experts per layer, but only activates 10 per token (plus one shared expert).

MoE architectures are also becoming increasingly popular among diffusion models. Many diffusion models, from text-to-image models such as SDXL~\citep{podell2023sdxlimprovinglatentdiffusion}, ERNIE-ViLG 2.0~\citep{feng2023ernievilg20improvingtexttoimage}, and eDIFF-I~\citep{balaji2023ediffitexttoimagediffusionmodels}, to recent video generation models such as Wan2.2~\citep{wan2025wanopenadvancedlargescale}, use different denoisers (as experts) for different denoising stages.

\paragraph{MoE Efficiency.} Several works have aimed to improve MoE efficiency via pruning, caching / prefetching, and offloading-aware serving.

On the pruning side,~\citep{muzio2024seermoesparseexpertefficiency} proposes a two-stage method that prunes the total number of experts using frequency count and then applies fine-tuning to recover accuracy.~\citep{lu2024expertsequalefficientexpert} minimizes a reconstruction loss and choose the subset of experts that best reconstructs the original layer's output and shows that this beats dropping least-used experts.~\citep{liu2024efficientexpertpruningsparse} introduces Efficient Expert Pruning (EEP) that employs gradient-free evolutionary strategy to prune and merge experts.~\citep{xie2024moeprunerpruningmixtureofexpertslarge} designs a novel pruning metric that incorporates MoE router weights information to identify and remove unimportant weights in expert layers. 

On the caching / prefetching and offloading-aware serving side,~\citep{xue2025moeinfinityefficientmoeinference} presents MoE-infinity that offloads certain experts to host memory and allows memory-constrained GPUs to serve MoE models. It uses expert activations to predict the set of experts to cache and prefetch.~\citep{song2025promoefastmoebasedllm} studies a similar setting and proposes ProMoE, which also uses activations to predict which experts will be needed soon and prefetch them.~\citep{tairin2025emoetaskawarememoryefficient} demonstrates that the set of experts that are needed across different layers and between similar prompts are correlated, and presents eMoE that predicts the set of experts to load onto GPU based on these observations in a task-aware manner.~\citep{zhang2025duoservemoedualphaseexpertprefetch} introduces DuoServe-MoE, which is an inference serving system that offloads certain expert weights to CPU and uses different scheduling for prefilling and decoding. Works in this thread generally do not study the tradeoff between latency cost and generation quality, but shares the similar goal of predicting the set of experts that will be used.

\paragraph{Options, s-MDPs, and hierarchical RL.} We cast expert-mask selection as a temporally extended control problem, where expert masks are options, switching the expert mask corresponds to option termination and can be regularized via an explicit switching latency cost.~\citep{SUTTON1999181} formalizes the options framework, showing that adding options to MDP induces a semi-MDP (s-MDP).~\citep{bacon2016optioncriticarchitecture} derives policy-gradient theorems for options and proposes the option-critic architecture for options learning.~\citep{harb2017waitingoptionlearning} points out that the options framework is the most useful when there's a deliberation cost between different options, where temporally extended structure naturally arises, and presents the variant of option-critic with deliberation cost.~\citep{klissarov2021flexibleoptionlearning} revisits intra-option learning in the context of deep reinforcement learning in order to enable updating
all options consistent with current primitive action choices, leading to better performance and data efficiency in option discovery.

Several recent papers are also utilizing the options framework or hierarchical RL to train language models. For example,~\citep{chai2025marlhfreinforcementlearninghuman} frames sequence of tokens as macro-actions and incorporates them into RLHF.~\citep{di2025enhancingmathreasoningsmallsized} proposes a hierarchical RL method based on GRPO for reasoning.~\citep{kobayashi2025emergenttemporalabstractionsautoregressive} proposes ``Internal RL,'' a hierarchical RL framework that improves sample efficiency in sparse-reward tasks by discovering and steering the temporally abstract action representations that naturally emerge within residual streams of pretrained autoregressive models.

%% file: implementation_details.tex
\section{Additional Implementation Details}
\label{sec:exp-setup}

\paragraph{Full algorithm.} We present the full training procedure in Algorithm~\ref{alg:moe_oc}.

\begin{algorithm}[!h]
\caption{MoE Option-Critic Training (Full)}
\label{alg:moe_oc}

\KwIn{MoE model with $L$ layers, $N$ experts per layer, top-$\activek$ routing; prompt dataset $\mathcal{D}$; teacher model $p_{\mathrm{teacher}}$; teacher mixing coefficient $\tau$; discount $\gamma$; GAE parameter $\lambda$; deliberation cost $\eta$, learning rates $\alpha_{\text{controller}}$, $\alpha_\text{intra}$. We use $\theta$ to denote the LLM parameters, $\nu$ to denote the termination head, $\psi$ to denote the critic parameters (including $V_\Omega$ and $Q_\Omega$), and $\phi$ to denote the option selection head parameters.}

\For{each training iteration}{
    Sample prompt $x \sim \mathcal{D}$\;

    \tcc{Rollout with teacher mixing}

    $\omega_0^{(\ell)} \leftarrow \mathrm{TopK}\bigl(g_0^{(\ell)},\, \maskk\bigr)$ for each layer $\ell$, where $g_0^{(\ell)} \in \mathbb{R}^N$ are the router logits at $t\!=\!0$\;

    \For{$t = 1, \ldots, T$}{
        \For{each layer $\ell$}{
            $\beta_t^{(\ell)} \leftarrow \sigma\left(\mathrm{MLP}_\beta\left(\mathrm{concat}\left(\overline{h}_t^{(\ell)}, \overline{z}^{(\ell)}(\omega_{t - 1}^{(\ell)})\right)\right)\right)$\;

            Sample $d_t^{(\ell)} \sim \mathrm{Bern}(\beta_t^{(\ell)})$\;

            $\omega_t^{(\ell)} \leftarrow \begin{cases} \mathrm{PL}\text{-}\mathrm{sample}(f_{\mathrm{sel}}(h_t^{(\ell)}),\, \maskk) & \text{if } d_t^{(\ell)}\!=\!1 \\ \omega_{t-1}^{(\ell)} & \text{otherwise}\end{cases}$\;

            Mask router to experts in $\omega_t^{(\ell)}$\;
        }

        $p_{\mathrm{mix}} \leftarrow (1 - \tau)\,\pi_{\omega,\theta}(\cdot \mid x, a_{<t}) + \tau\, p_{\mathrm{teacher}}(\cdot \mid x, a_{<t})$\;

        Sample $a_t \sim p_{\mathrm{mix}}$\;

        $w_t \leftarrow \pi_{\omega,\theta}(a_t) / p_{\mathrm{mix}}(a_t)$\;

        $r_t \leftarrow \log p_{\mathrm{teacher}}(a_t \mid x, a_{<t}) - \log \pi_{\omega, \theta}(a_t \mid x, a_{<t})$\;
    }
}
\end{algorithm}

\begin{algorithm}[!h]
\textbf{Algorithm~\ref{alg:moe_oc} continued.}\;

\For{same training iteration}{
    \tcc{Per-layer critic and termination gradient computation}

    \For{each layer $\ell$}{
        \For(\tcp*[f]{GAE($\lambda$)}){$t = T, \ldots, 1$}{
            $\beta_{t + 1}^{(\ell)} \leftarrow \sigma\left(\mathrm{MLP}_\beta\left(\mathrm{concat}\left(\overline{h}_{t + 1}^{(\ell)}, \overline{z}^{(\ell)}(\omega_{t}^{(\ell)})\right)\right)\right)$\;

            $U_t \leftarrow \beta_{t + 1}^{(\ell)}\, V_\Omega(h_{t+1}^{(\ell)}) + (1 - \beta_{t + 1}^{(\ell)})\, Q_\Omega(h_{t+1}^{(\ell)}, \omega_t^{(\ell)})$\;

            $\delta_t^V \leftarrow r_t + \gamma\, V_\Omega(h_{t+1}^{(\ell)}) - V_\Omega(h_t^{(\ell)})$\;

            $\hat{A}_t^V \leftarrow \delta_t^V + \gamma\lambda\,\hat{A}_{t+1}^V$\;

            $\delta_t^Q \leftarrow r_t + \gamma\, U_t - Q_\Omega(h_t^{(\ell)}, \omega_t^{(\ell)})$\;

            $\hat{A}_t^Q \leftarrow \delta_t^Q + \gamma\lambda\,\hat{A}_{t+1}^Q$\;
        }

        \For{$t = 1, \ldots, T$}{
            $\hat{V}_t^{\mathrm{targ}} \leftarrow V_\Omega(h_t^{(\ell)}) + \hat{A}_t^V$; \enspace $\hat{Q}_t^{\mathrm{targ}} \leftarrow Q_\Omega(h_t^{(\ell)}, \omega_t^{(\ell)}) + \hat{A}_t^Q$\;

            $d\nu \mathrel{-\!=} w_t \cdot \nabla_{\nu} \beta(h_t^{(\ell)}, \omega_{t - 1}^{(\ell)}) \cdot \big(Q_\Omega(h_t^{(\ell)}, \omega_{t-1}^{(\ell)}) - V_\Omega(h_t^{(\ell)}) + \eta\big)$\;

            \If{$d_t^{(\ell)} = 1$}{
                $d\phi \mathrel{+\!=} w_t \cdot \nabla_\phi \log \pi_{\mathrm{sel}}(\omega_t^{(\ell)} \mid h_t^{(\ell)}) \cdot \big(Q_\Omega(h_t^{(\ell)}, \omega_t^{(\ell)}) - V_\Omega(h_t^{(\ell)})\big)$\;
            }

            $d\psi \mathrel{-\!=} \nabla_{\psi}\Big[\big(V_\Omega(h_t^{(\ell)}) - \hat{V}_t^{\mathrm{targ}}\big)^2 + \big(Q_\Omega(h_t^{(\ell)}, \omega_t^{(\ell)}) - \hat{Q}_t^{\mathrm{targ}}\big)^2\Big]$\;
        }
    }

    \tcc{Intra-option policy gradient computation}

    \For{$t = 1, \ldots, T$}{
        $\bar{G}_t \leftarrow \sum_{j \ge 0}\gamma^j r_{t+j}$\;

        $d\theta \mathrel{+\!=} w_t \cdot \nabla_\theta \log\pi_{\omega,\theta}(a_t \mid x, a_{<t}) \cdot \bar{G}_t$\;
    }

    $(\nu, \psi, \phi) \leftarrow (\nu, \psi, \phi) + \alpha_\text{controller} \cdot \frac{1}{L}\, (d\nu, d\psi, d\phi)$; \enspace $\theta \leftarrow \theta + \alpha_\text{intra} \cdot d\theta$\;
}
\end{algorithm}

\paragraph{Controller architecture details.} Each MoE layer has an independent controller. The DeepSets expert set encoder uses a learned embedding dimension of $d_e = 128$ and a two-layer MLP with GELU activation and hidden dimension $1024$. The termination head, option-value head $Q_\Omega$, and expert selection head all use hidden dimension $1024$. The state-value head $V_\Omega$ is a single linear layer initialized from the router weights. The termination head's bias is initialized to $-3$ (corresponding to an initial switch probability of $\sigma(-3) \approx 0.05$, encouraging temporal continuity from the beginning). RMSNorm is applied to balance the scale of $h_t^{(\ell)}$ and $z_t^{(\ell)}$ before concatenation.

\paragraph{Advantage normalization} For the termination gradient, the raw advantage is
$$A_t^{\mathrm{term},(\ell)} = Q_\Omega(h_t^{(\ell)}, \omega_{t-1}^{(\ell)}) - V_\Omega(h_t^{(\ell)}) + \eta.$$
We apply RMS normalization (without mean centering) independently within each layer $\ell$ over all timesteps $t > 0$:
$$\hat{A}_t^{\mathrm{term},(\ell)} = \frac{A_t^{\mathrm{term},(\ell)}}{\mathrm{RMS}(A^{\mathrm{term},(\ell)})}, \quad \text{where} \quad \mathrm{RMS}(A^{\text{term}, (\ell)}) = \sqrt{\frac{1}{T - 1}\sum_{t = 1}^{T - 1}\left(A_t^{\mathrm{term},(\ell)}\right)^2}.$$
This preserves the sign of advantages while stabilizing scale, which we found to be important for the stability of training.

When a switch occurs ($d_t^{(\ell)} = 1$), the per-layer option selection advantage is 
$$A_t^{\mathrm{sel},(\ell)} = Q_\Omega(h_t^{(\ell)}, \omega_t^{(\ell)}) - V_\Omega(h_t^{(\ell)}).$$ 
We apply the same RMS normalization, computed over switch positions only:
$$
\hat{A}_t^{\mathrm{sel, (\ell)}} = \frac{A_t^{\mathrm{sel}, (\ell)}}{\mathrm{RMS}(A^{\mathrm{sel}, (\ell)})}, \quad \text{where} \quad \mathrm{RMS}(A^{\text{sel}, (\ell)}) = \sqrt{\frac{1}{|\mathcal{S^{(\ell)}}|}\sum_{t \in \mathcal{S}^{(\ell)}} (A_t^{\mathrm{sel}, (\ell)})^2},
$$
and $\mathcal{S}^{(\ell)} = \{t : d_t^{(\ell)} = 1, t > 0\}$ is the set of switch positions at layer $\ell$.

The intra-option policy gradient (Theorem~\ref{thm:intra-option}) requires $Q_U(s, \omega, a)$, the value of taking action $a$ under state-option pair $(s, \omega)$. We estimate this with the Monte Carlo return:
$$
\bar{G}_t = \sum_{j=0}^{T-t-1} \gamma^j r_{t+j},
$$
which serves as an unbiased estimate of $Q_U(s_t, \omega_t, a_t)$. We use $\bar{G}_t$ directly as the advantage (without subtracting a baseline as in~\citep{harb2017waitingoptionlearning}, which we found to work better empirically). Before applying the policy gradient, we standardize $\bar{G}_t$ across all $T$ response tokens:
$$
\hat{A}_t^{\mathrm{intra}} = \frac{\bar{G}_t - \mu}{\sigma}, \quad \text{where} \quad \mu = \frac{1}{T}\sum_{t=1}^{T} \bar{G}_t, \quad \sigma = \sqrt{\frac{1}{T}\sum_{t=1}^{T} (\bar{G}_t - \mu)^2}.
$$
This advantage is shared across all layers.

\paragraph{Implementation detail of reconstruction loss minimization.} For reconstruction loss minimization,~\citep{lu2024expertsequalefficientexpert} prunes out a set of experts so that the remaining set of experts minimizes a reconstruction loss. For each MoE layer $\ell$, we cache input-output pairs $(x_i, \mathcal{F}(x_i))$ by running a forward pass over the calibration data, where $\mathcal{F}$ denotes the full MoE layer. Given a candidate subset $\mathbf{C} \subseteq \{1, \ldots, N\}$ with $|\mathbf{C}| = \maskk$, the pruned layer $\mathcal{F}'(\cdot, \mathbf{C})$ masks all experts not in $\mathbf{C}$, recomputes the top-$\activek$ routing and softmax normalization over only the allowed experts, and produces a weighted combination of expert outputs. The objective is
$$
\min_{\mathbf{C}: |\mathbf{C}| = \maskk} \left\| \mathcal{F}'(X, \mathbf{C}) - \mathcal{F}(X) \right\|_F,
$$
where $X$ denotes the cached calibration inputs and $\|\cdot\|_F$ is the Frobenius norm. The original method in~\citep{lu2024expertsequalefficientexpert} solves this via exhaustive enumeration over all $\binom{N}{\maskk}$ subsets, which is feasible in their setting where $N = 8$. In our setting with $N = 32$ experts, exhaustive search is computationally infeasible. We therefore employ a greedy forward selection procedure: starting from an empty set, we iteratively add the expert that yields the largest reduction in reconstruction loss, repeating $\maskk$ times. This compromise is due to the inherent limitation of the scalability of their method, and our implementation is our best effort to map their method in our setting.

%% file: additional_experiment.tex
\section{Additional Experimental Results}
\label{sec:additional_experiment}
In this section, we present additional experimental results and analysis. 

\begin{figure*}[!h]
    \centering
    \begin{subfigure}[t]{0.49\linewidth}
        \centering
        \includegraphics[width=\linewidth]{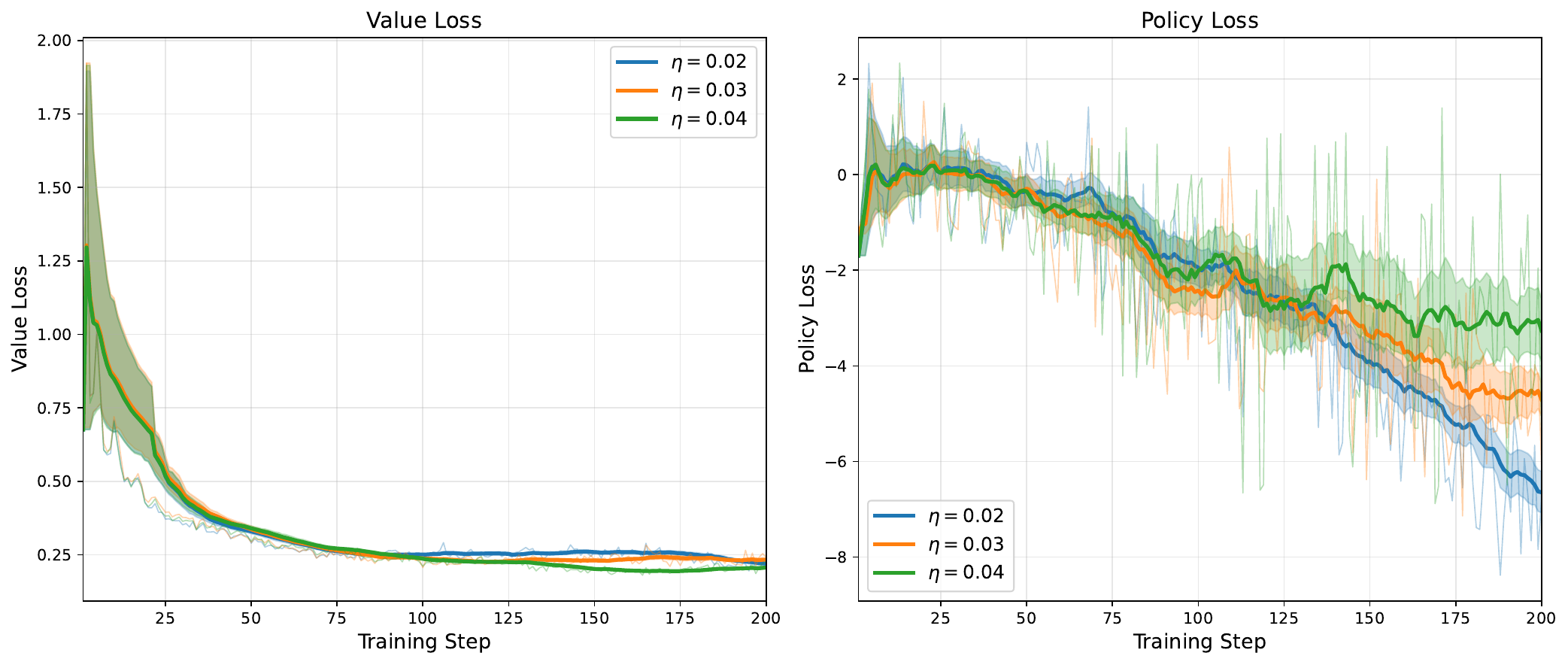}
        \caption{Loss curves for $\maskk = 8$.}
        \label{fig:loss_exp8}
    \end{subfigure}\hfill
    \begin{subfigure}[t]{0.49\linewidth}
        \centering
        \includegraphics[width=\linewidth]{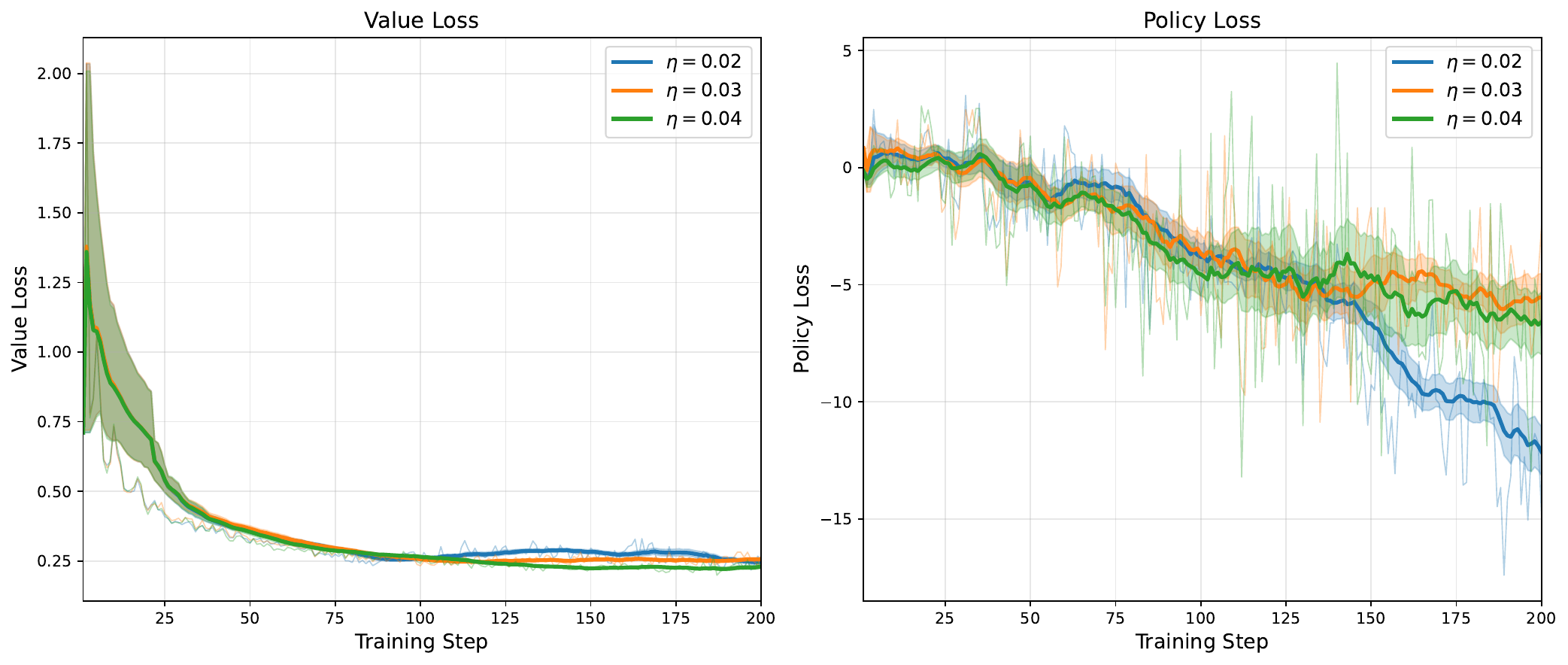}
        \caption{Loss curves for $\maskk = 16$}
        \label{fig:loss_exp16}
    \end{subfigure}
    \caption{Loss curves for $\maskk = 8$ and $\maskk = 16$, respectively. Running averages and confidence bands plotted the same way as in Figure~\ref{fig:training_curve}.}
    \label{fig:loss_curve}
\end{figure*}

\paragraph{Loss curves.} We present the loss curves in Figure~\ref{fig:loss_curve}. We can see that the value loss steadily decreases, which indicates that the value and option-value heads are learning the value of states and state-option pairs effectively.

\begin{figure*}[!h]
    \centering
    \begin{subfigure}[t]{0.49\linewidth}
        \centering
        \includegraphics[width=\linewidth]{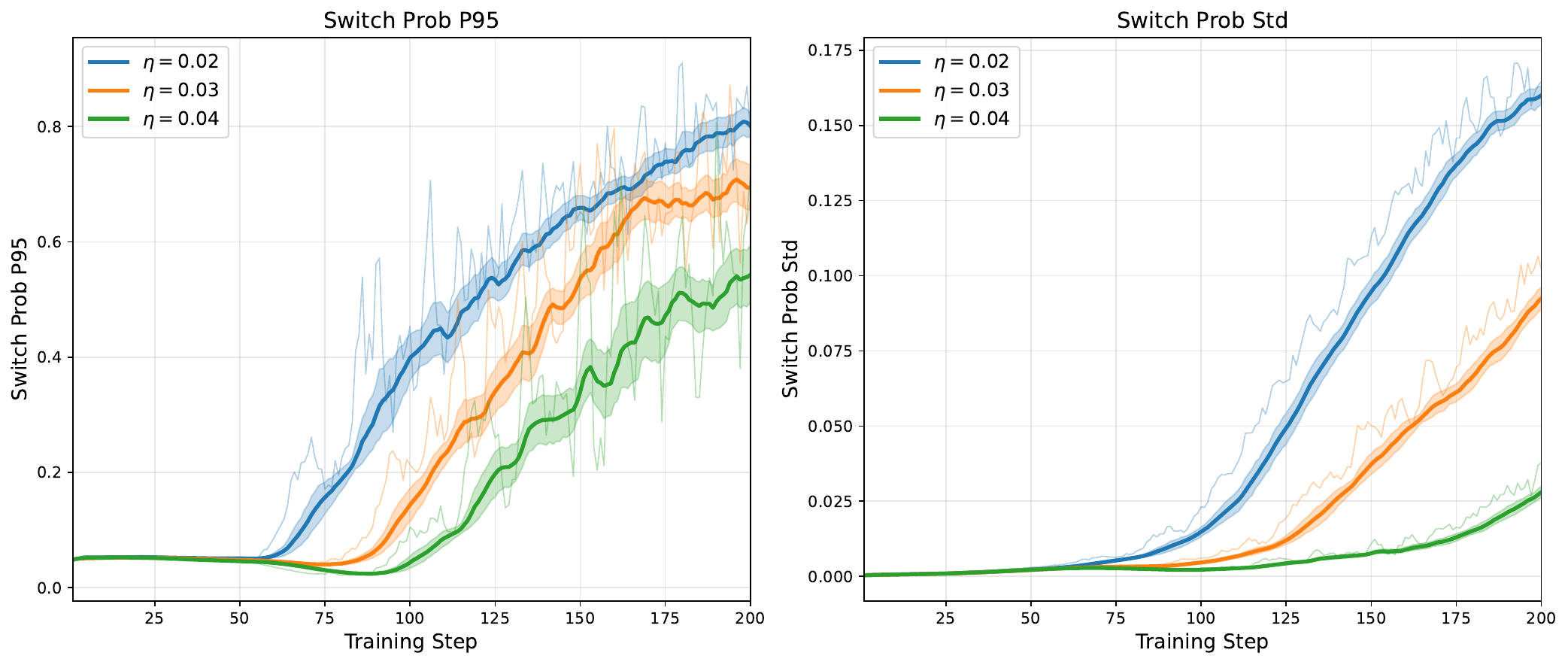}
        \caption{Switch rate curves for $\maskk = 8$.}
        \label{fig:switch_exp8}
    \end{subfigure}\hfill
    \begin{subfigure}[t]{0.49\linewidth}
        \centering
        \includegraphics[width=\linewidth]{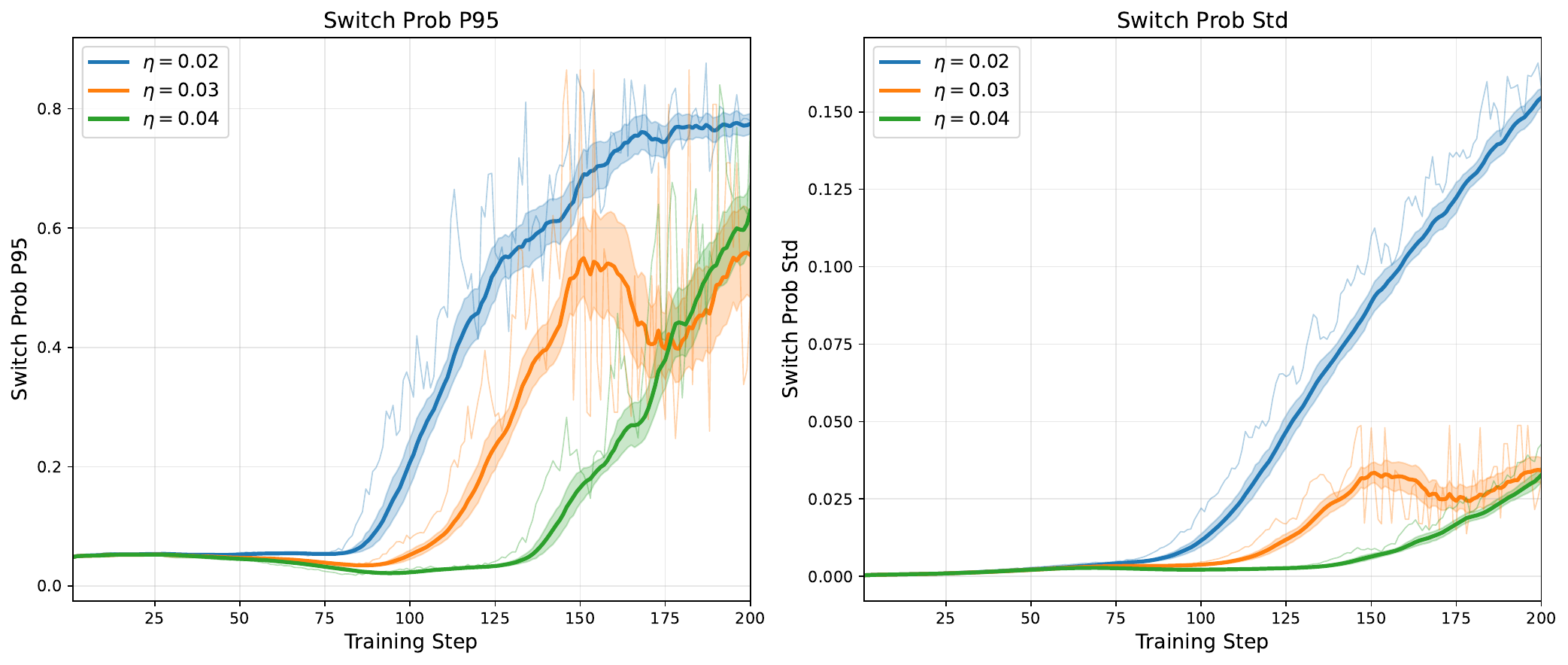}
        \caption{Switch rate curves for $\maskk = 16$.}
        \label{fig:switch_exp16}
    \end{subfigure}
    \caption{Additional switch rate curves for $\maskk = 8$ and $\maskk = 16$, respectively. Running averages and confidence bands plotted the same way as in Figure~\ref{fig:training_curve}.}
    \label{fig:switch_curve}
\end{figure*}

\paragraph{Additional switch rate curves.} We present additional switch rate curves in Figure~\ref{fig:switch_curve}. For each subplot, the plot on the left presents the switch probability at 95 percentile over the trajectory throughout training, while the plot on the right presents the standard deviation of the switch probabilities over the trajectory throughout training. We can see that though switch rate first decreases and then slightly increases (as shown in Figure~\ref{fig:training_curve}), both the 95 percentile switch rate and the standard deviation of the switch rates steadily increases, indicating that the termination head is learning to distinguish when to switch and when to not switch effectively.

\begin{figure*}[!h]
    \centering
    \begin{subfigure}[t]{0.49\linewidth}
        \centering
        \includegraphics[width=\linewidth]{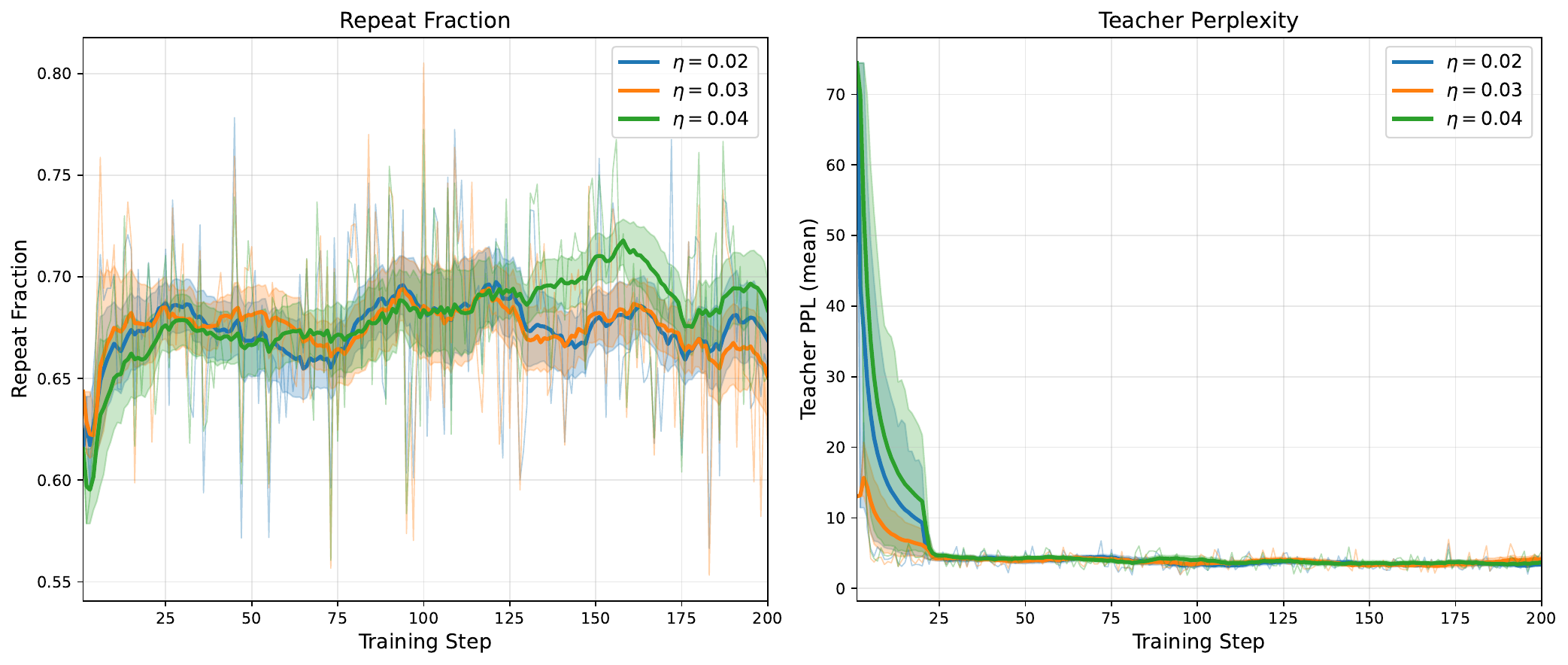}
        \caption{Repetition rate and perplexity curves for $\maskk = 8$.}
        \label{fig:ppl_exp8}
    \end{subfigure}\hfill
    \begin{subfigure}[t]{0.49\linewidth}
        \centering
        \includegraphics[width=\linewidth]{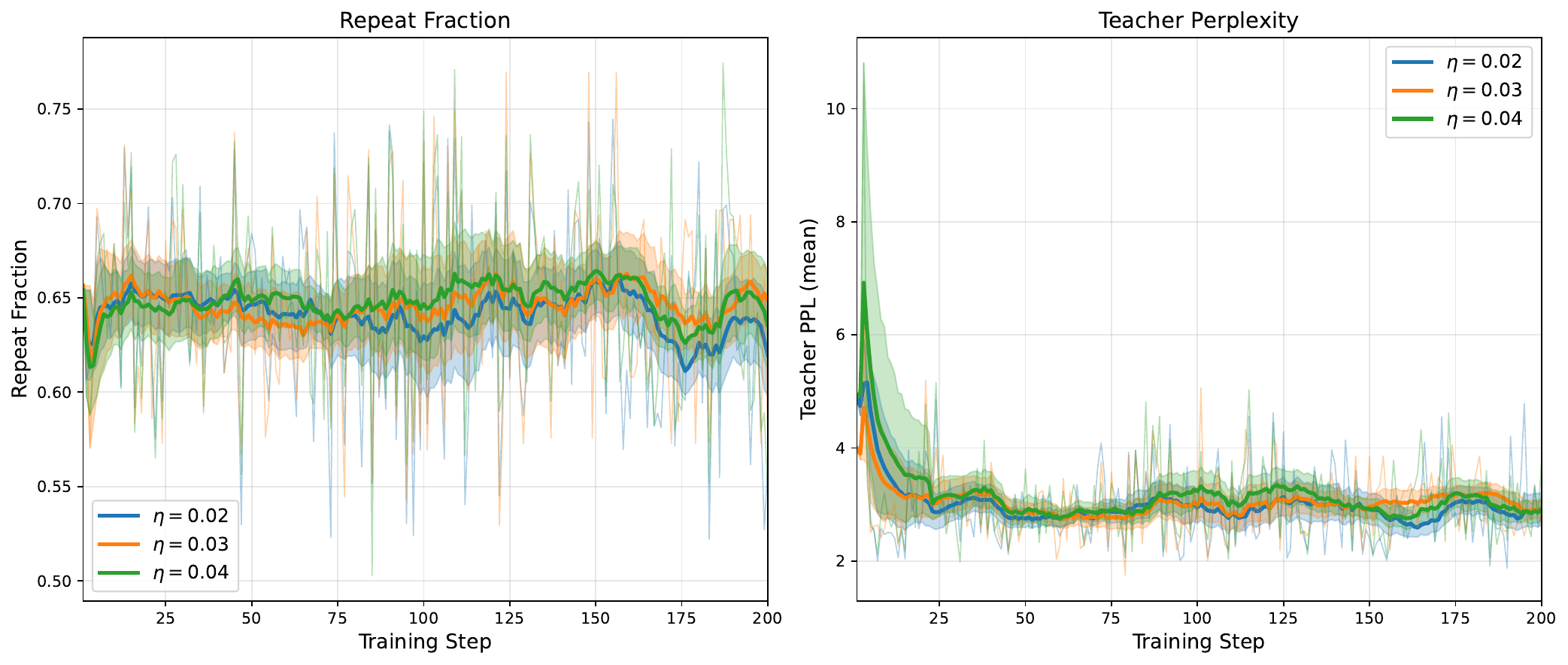}
        \caption{Repetition rate and perplexity curves for $\maskk = 16$.}
        \label{fig:ppl_exp16}
    \end{subfigure}
    \caption{Repetition rate and perplexity curves for $\maskk = 8$ and $\maskk = 16$, respectively. Running averages and confidence bands plotted the same way as in Figure~\ref{fig:training_curve}.}
    \label{fig:ppl_curve}
\end{figure*}

\paragraph{Training stability.} A common failure mode of MoE models under constrained routing is catastrophic repetition, where the model degenerates into producing repetitive content (concrete examples in Section~\ref{sec:examples}). Figure~\ref{fig:ppl_curve} shows that our method avoids this failure mode: the repetition rate ($1 -\text{fraction of unique tokens per trajectory}$) remains stable within a healthy range throughout training for both $\maskk = 8$ and $\maskk = 16$. The right panels show the perplexity of the frozen teacher model (i.e., the original \texttt{gpt-oss-20b} without any controllers or weight updates) evaluated on the student's generated trajectories, which decreases over training, indicating that the student's outputs become more aligned with the teacher rather than diverging.

\begin{figure*}[!h]
    \centering
    \begin{subfigure}[t]{0.32\linewidth}
        \centering
        \includegraphics[width=.8\linewidth]{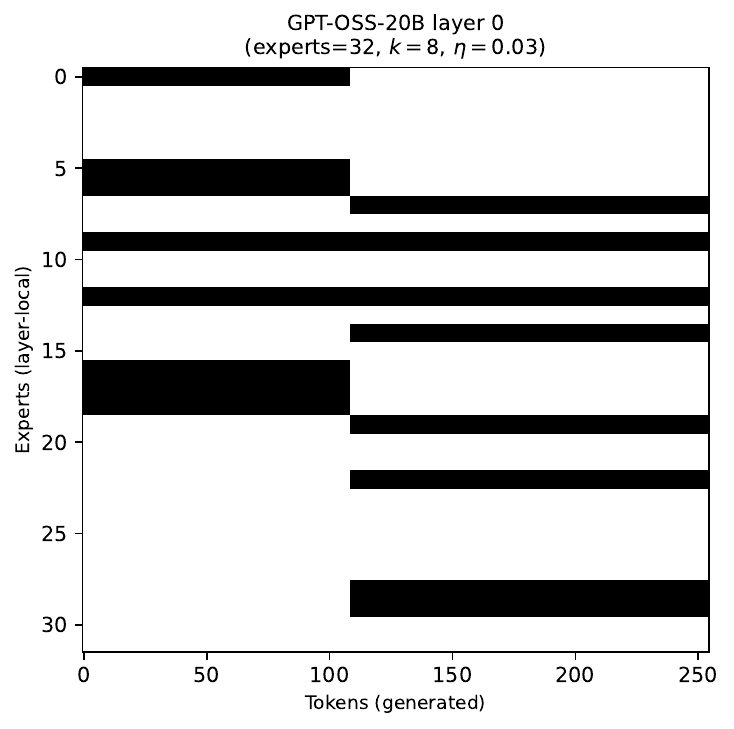}
        \caption{Layer 0}
        \label{fig:k8eta0.03layer0}
    \end{subfigure}\hfill
    \begin{subfigure}[t]{0.32\linewidth}
        \centering
        \includegraphics[width=.8\linewidth]{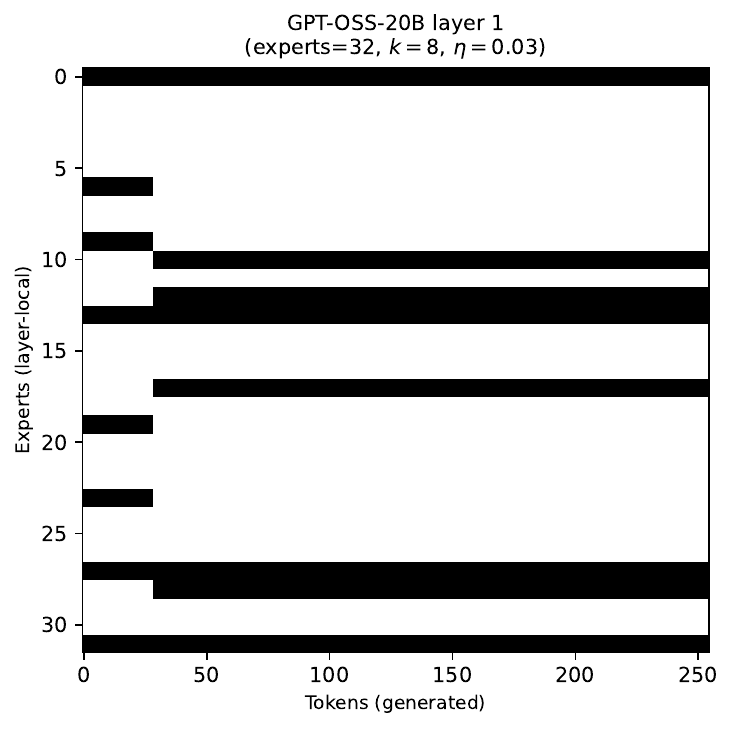}
        \caption{Layer 1}
        \label{fig:k8eta0.03layer1}
    \end{subfigure}\hfill
    \begin{subfigure}[t]{0.32\linewidth}
        \centering
        \includegraphics[width=.8\linewidth]{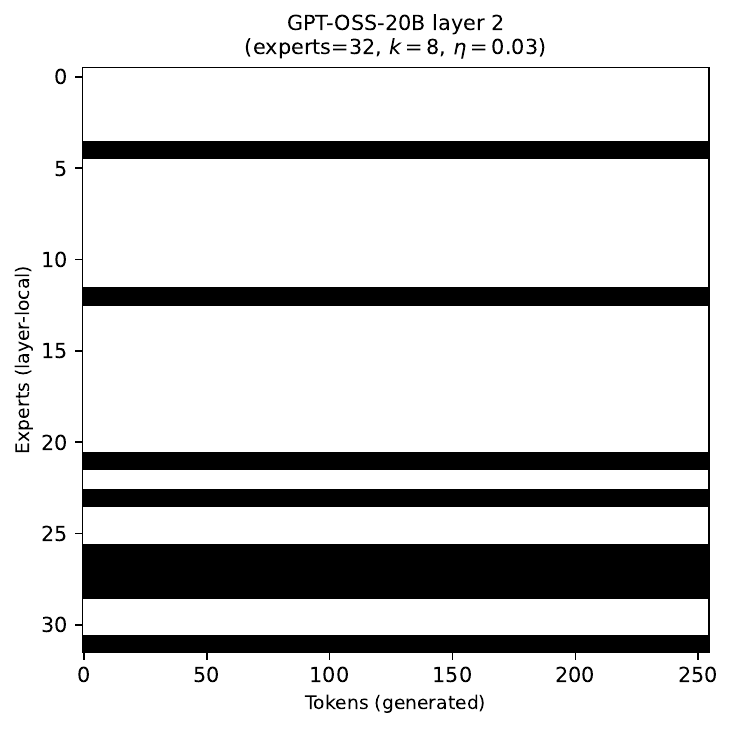}
        \caption{Layer 2}
        \label{fig:k8eta0.03layer2}
    \end{subfigure}
    \caption{Options in layer 0, 1, 2 throughout the trajectory with gpt-oss-20b, $\maskk = 8, \eta = 0.03$.}
    \label{fig:temporal_continuity_k8_eta0.03}
\end{figure*}

\begin{figure*}[!h]
    \centering
    \begin{subfigure}[t]{0.32\linewidth}
        \centering
        \includegraphics[width=.8\linewidth]{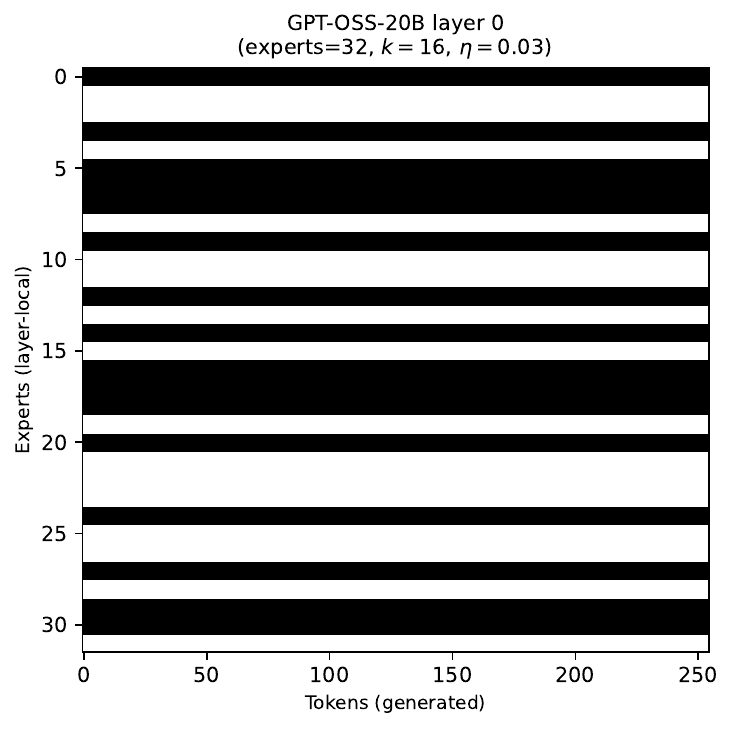}
        \caption{Layer 0}
        \label{fig:k16eta0.03layer0}
    \end{subfigure}\hfill
    \begin{subfigure}[t]{0.32\linewidth}
        \centering
        \includegraphics[width=.8\linewidth]{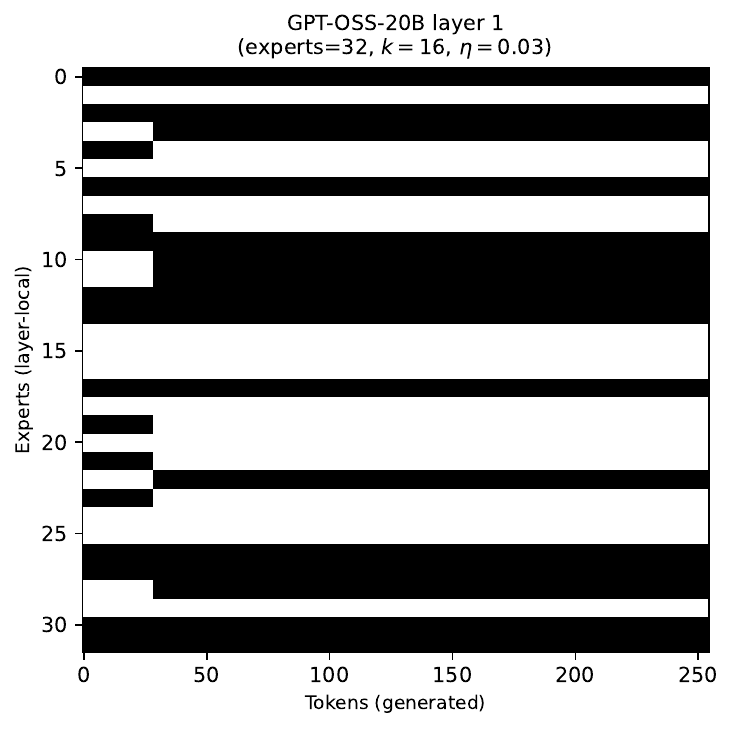}
        \caption{Layer 1}
        \label{fig:k16eta0.03layer1}
    \end{subfigure}\hfill
    \begin{subfigure}[t]{0.32\linewidth}
        \centering
        \includegraphics[width=.8\linewidth]{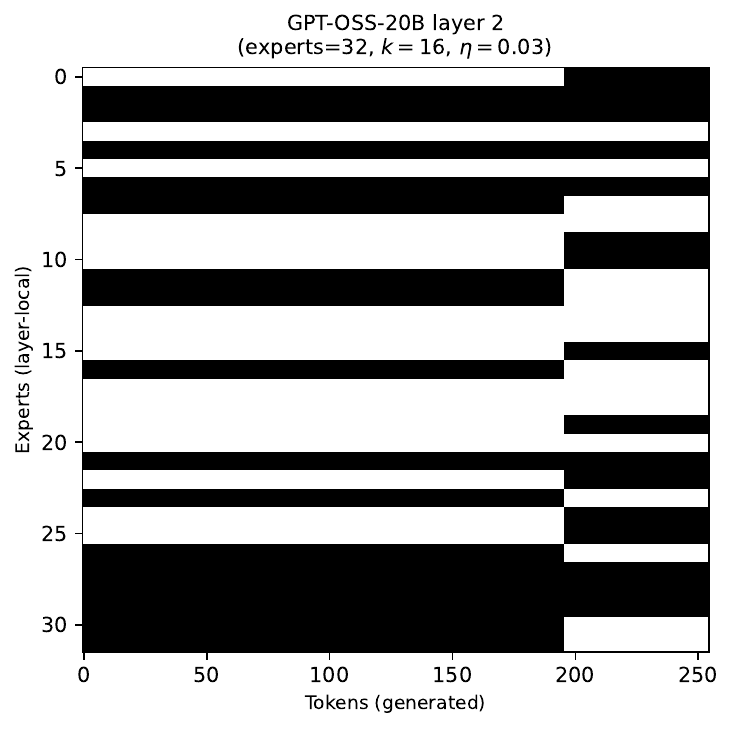}
        \caption{Layer 2}
        \label{fig:k16eta0.03layer2}
    \end{subfigure}
    \caption{Options in layer 0, 1, 2 throughout the trajectory with gpt-oss-20b, $\maskk = 16, \eta = 0.03$.}
    \label{fig:temporal_continuity_k16_eta0.03}
\end{figure*}

\begin{figure*}[!h]
    \centering
    \begin{subfigure}[t]{0.32\linewidth}
        \centering
        \includegraphics[width=.8\linewidth]{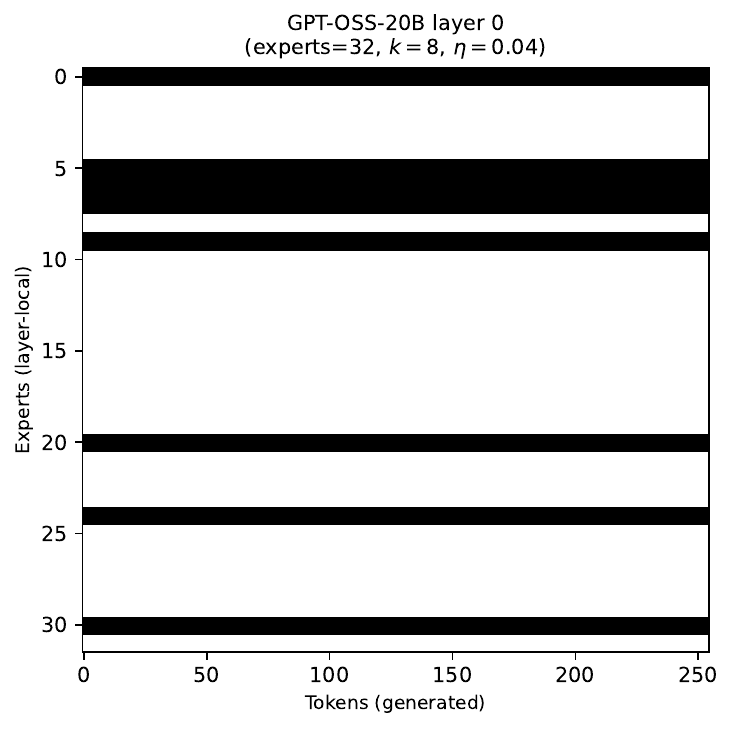}
        \caption{Layer 0}
        \label{fig:k8eta0.04layer0}
    \end{subfigure}\hfill
    \begin{subfigure}[t]{0.32\linewidth}
        \centering
        \includegraphics[width=.8\linewidth]{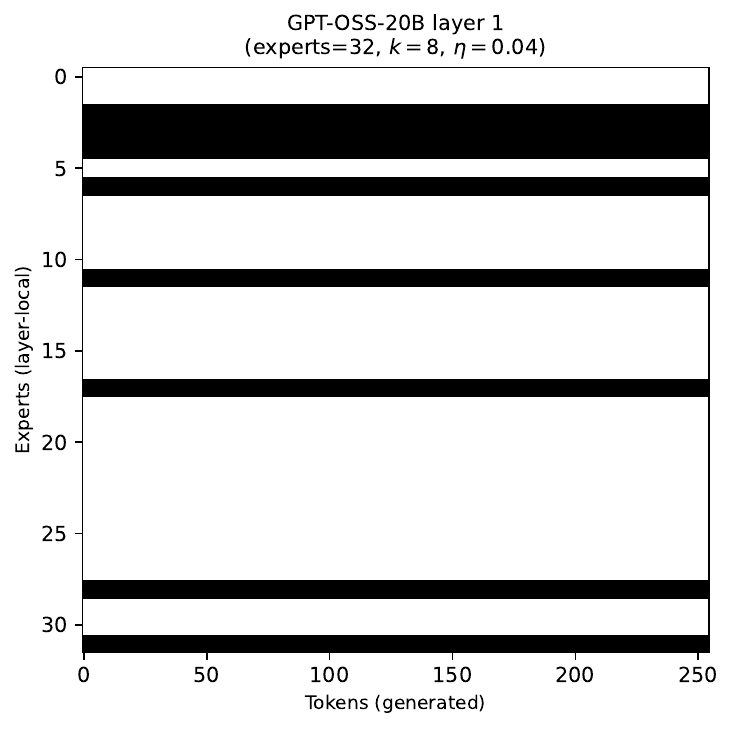}
        \caption{Layer 1}
        \label{fig:k8eta0.04layer1}
    \end{subfigure}\hfill
    \begin{subfigure}[t]{0.32\linewidth}
        \centering
        \includegraphics[width=.8\linewidth]{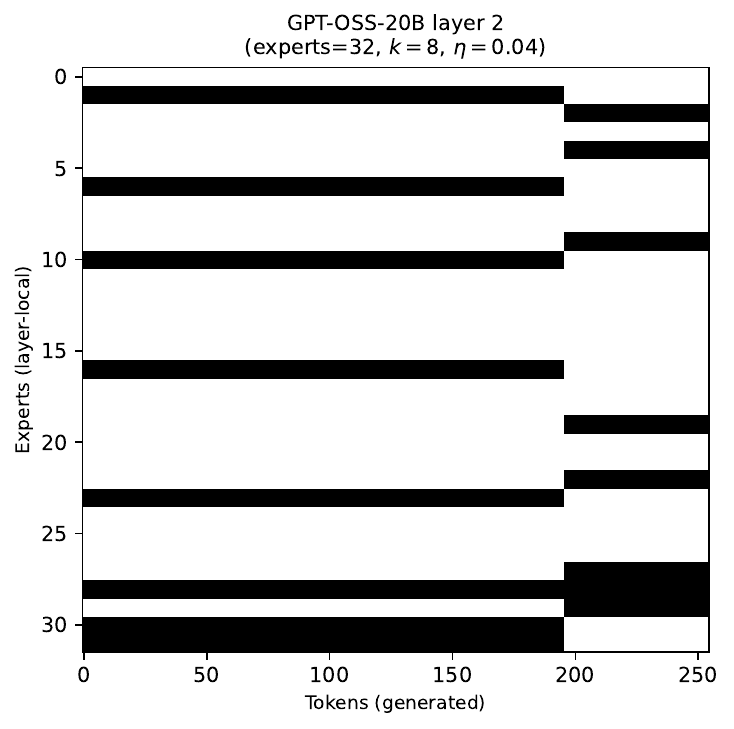}
        \caption{Layer 2}
        \label{fig:k8eta0.04layer2}
    \end{subfigure}
    \caption{Options in layer 0, 1, 2 throughout the trajectory with gpt-oss-20b, $\maskk = 8, \eta = 0.04$.}
    \label{fig:temporal_continuity_k8_eta0.04}
\end{figure*}

\begin{figure*}[!h]
    \centering
    \begin{subfigure}[t]{0.32\linewidth}
        \centering
        \includegraphics[width=.8\linewidth]{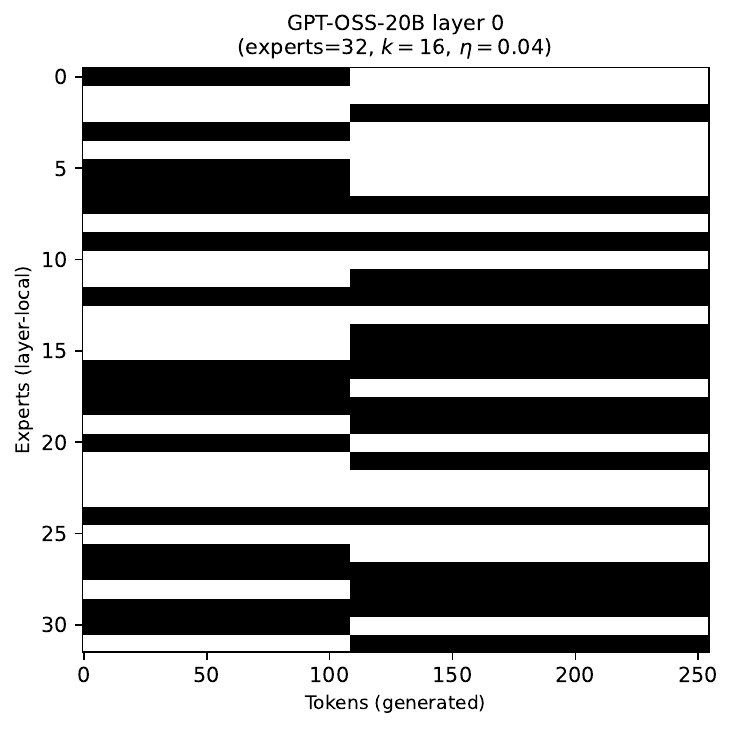}
        \caption{Layer 0}
        \label{fig:k16eta0.04layer0}
    \end{subfigure}\hfill
    \begin{subfigure}[t]{0.32\linewidth}
        \centering
        \includegraphics[width=.8\linewidth]{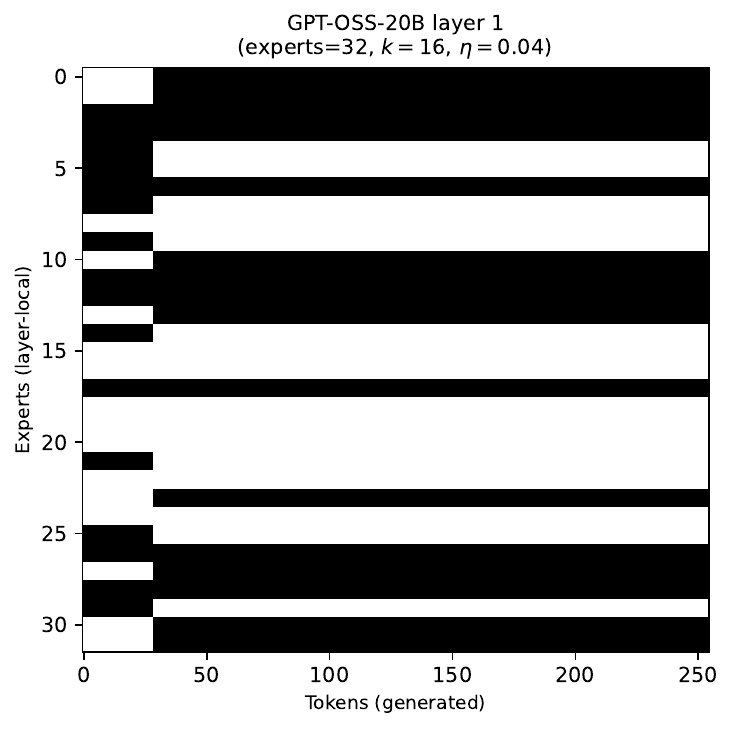}
        \caption{Layer 1}
        \label{fig:k16eta0.04layer1}
    \end{subfigure}\hfill
    \begin{subfigure}[t]{0.32\linewidth}
        \centering
        \includegraphics[width=.8\linewidth]{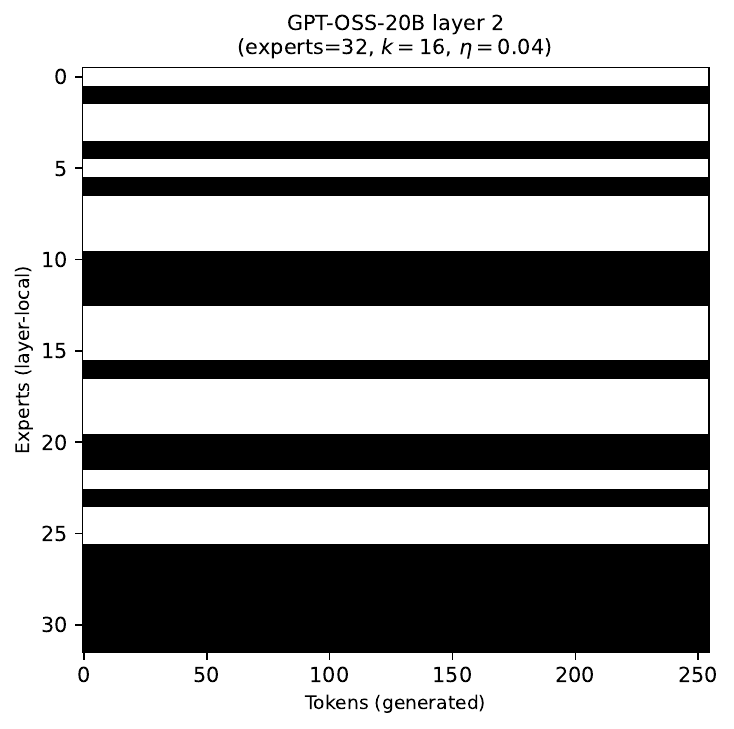}
        \caption{Layer 2}
        \label{fig:k16eta0.04layer2}
    \end{subfigure}
    \caption{Options in layer 0, 1, 2 throughout the trajectory with gpt-oss-20b, $\maskk = 16, \eta = 0.04$.}
    \label{fig:temporal_continuity_k16_eta0.04}
\end{figure*}

\paragraph{Temporal continuity with controller.} We present additional plots demonstrating temporal continuity of options under our trained controller using the same setup as in Section~\ref{sec:exp}. Figure~\ref{fig:temporal_continuity_k8_eta0.03} and~\ref{fig:temporal_continuity_k16_eta0.03} plot the active options throughout the trajectory for $\eta = 0.03$, while Figure~\ref{fig:temporal_continuity_k8_eta0.04} and Figure~\ref{fig:temporal_continuity_k16_eta0.04} plot for $\eta = 0.04$. We can see that options almost always demonstrate significant temporal continuity across different layers and different deliberation costs.

%% file: limitation.tex
\section{Limitations and Future Directions}
\label{sec:limitations}
Our work proposes temporally extended MoE as a design philosophy and demonstrates its feasibility through a principled option-critic framework. We discuss several limitations and promising directions for future research.

\paragraph{From philosophy to deployment.}
While we identify three concrete opportunities enabled by temporal extension --- memory-efficient inference, chunk-wise training, and continual learning with expandable experts --- our current experiments focus on validating that temporally extended routing can be learned with graceful degradation in generation quality. Building end-to-end systems that realize these memory savings is substantial systems work that we leave to future research. Similarly, the deliberation cost in our framework is a tunable hyperparameter rather than a measured hardware latency; grounding it in actual expert loading times for a specific deployment target would make the cost-quality tradeoff directly actionable.

\paragraph{Realizing temporal continuity in pre-training.}
Our framework does post-training on a pre-trained MoE model. An intriguing alternative is to encode temporal continuity directly into the pre-training objective. As GPU memory becomes an increasingly binding constraint for scaling, building temporal structure into routing from the start could yield models that are inherently more memory-friendly without requiring a separate controller. Moreover, natural language itself exhibits rich temporal structure (e.g., topics, arguments, or reasoning chains that persist over extended spans within a trajectory) and there may be richer connections between temporally abstract representations in language and temporally extended expert routing that future work can exploit.

\paragraph{Per-layer vs.\ cross-layer options.}
In our framework, each layer maintains its own independent option, meaning that expert masks may switch at different token positions across layers. In practice, the most straightforward way to realize the memory benefits discussed in Section~\ref{sec:motivation} would be to switch expert masks across all layers simultaneously, so that a single offload/onload event can swap the entire active expert set at once. This would also simplify chunk-wise training, where a chunk boundary aligned across all layers naturally defines the segments for memory-efficient forward-backward passes. However, defining options as joint expert masks across all $L$ layers would result in a combinatorially larger option space, making the learning problem significantly harder. Our per-layer formulation is a practical compromise that makes learning tractable while still demonstrating the benefits of temporal extension. Bridging this gap is an important direction for future work.

\paragraph{Evaluation scope.} Our evaluation covers three benchmarks that span distinct capabilities: mathematical reasoning (MATH), broad knowledge (MMLU), and multilingual understanding (MMMLU). However, we have not yet included other important dimensions such as code generation, long-form instruction following, or open-ended conversation. In addition, we evaluate on 200 randomly selected questions per benchmark rather than the full test sets, which is sufficient to establish statistically significant comparisons (as reflected in the confidence intervals in Tables~\ref{tab:results-16experts} and~\ref{tab:results-8experts}) but may not fully capture performance variation across fine-grained subtopics. A more comprehensive evaluation across a wider range of benchmarks and larger sample sizes would strengthen the empirical picture.

\paragraph{Disentangling temporal extension from self-distillation.}
Our method combines two sources of improvement over static pruning baselines: (1) the ability to dynamically switch expert masks across tokens, and (2) on-policy self-distillation that adapts the model's weights to better match the base model. Note that the latter is not an ad hoc addition, but an inherent component of the option-critic framework. Drawing this connection between between intra-option learning in option-critic framework and on-policy distillation is an important part of our contribution. Nonetheless, since our primary baselines~\citep{lu2024expertsequalefficientexpert} are training-free post-training methods, one might ask whether the gains we observe stem primarily from temporal extension or from the additional parameter updates. To fully disentangle these two factors, a natural ablation would be to apply the same on-policy distillation objective to a statically pruned model with a fixed expert mask, isolating the effect of self-distillation alone. We leave this ablation to future work and note that it would provide a more fine-grained understanding of the relative contributions of dynamic routing and weight adaptation.

%% file: example.tex
\section{Examples}
\label{sec:examples}
In this section, we present some concrete rollouts of different methods compared in the paper. We look at a number theory problem in MATH dataset, using $\maskk = 16$ experts for each method. We set token sampling temperature to $0.5$ and max token count to $2048$. For our method, we use a deliberation cost of $0.02$ and take the checkpoint at $120^{\text{th}}$ step. While all methods are capable of producing coherent outputs to different extents, this particular example illustrates a case where only our method maintains coherent reasoning, whereas the baseline methods degrade to varying degrees of gibberish.


\promptblock{
Problem: The least common multiple of $1!+2!$, $2!+3!$, $3!+4!$, $4!+5!$, $5!+6!$, $6!+7!$, $7!+8!$, and $8!+9!$ can be expressed in the form $a\cdot b!$, where $a$ and $b$ are integers and $b$ is as large as possible. What is $a+b$?
}

\responseblock{
(Ours) We need to find the LCM of numbers: $n_i = i! + (i+1)!$ for $i$ from $1$ to $8$ inclusive. \vspace{1em}

Compute each expression: $i! + (i+1)! = i! + (i+1) \cdot i! = (1 + (i+1)) i! = (i+2) i!$. \vspace{1em}

So $n_i = (i+2) \cdot i!$ for $i=1 \dots 8$. \vspace{1em}

Thus $n_i = (i+2) \cdot i!$. \vspace{1em}

We need the LCM of these numbers. \vspace{1em}

We can express each as $(i+2) \cdot i!$. Let's find prime factorizations. \vspace{1em}

But maybe we can compute the LCM using factorials. \vspace{1em}

Observe that $i!$ divides $(i+1)!$, etc. But we have an $(i+2)$ factor. \vspace{1em}

We need the LCM of numbers of the form $(i+2) \cdot i!$. \vspace{1em}

We can factor each as $(i+2) \cdot i!$. \vspace{1em}

We can compute the LCM by considering prime powers. \vspace{1em}

Let's compute for each $i$: \vspace{1em}

$i=1$: $1! + 2! = 1 + 2 = 3$. Our formula: $(1+2) \cdot 1! = 3 \cdot 1 = 3$. \vspace{1em}

$i=2$: $2! + 3! = 2 + 6 = 8$. Formula: $(2+2) \cdot 2! = 4 \cdot 2 = 8$. \vspace{1em}

$i=3$: $3! + 4! = 6 + 24 = 30$. Formula: $(3+2) \cdot 3! = 5 \cdot 6 = 30$. \vspace{1em}

$i=4$: $4! + 5! = 24 + 120 = 144$. Formula: $(4+2) \cdot 4! = 6 \cdot 24 = 144$. \vspace{1em}

$i=5$: $5! + 6! = 120 + 720 = 840$. Formula: $(5+2) \cdot 5! = 7 \cdot 120 = 840$. \vspace{1em}

$i=6$: $6! + 7! = 720 + 5040 = 5760$. Formula: $(6+2) \cdot 6! = 8 \cdot 720 = 5760$. \vspace{1em}

$i=7$: $7! + 8! = 5040 + 40320 = 45360$. Formula: $(7+2) \cdot 7! = 9 \cdot 5040 = 45360$. \vspace{1em}

$i=8$: $8! + 9! = 40320 + 362880 = 403200$. Formula: $(8+2) \cdot 8! = 10 \cdot 40320 = 403200$. \vspace{1em}

Thus the numbers are: $3, 8, 30, 144, 840, 5760, 45360, 403200$. \vspace{1em}

We need the LCM of all these numbers. \vspace{1em}

We can compute the LCM by factoring each. \vspace{1em}

Let's factor each: \vspace{1em}

$3$: prime $3$. \vspace{1em}

$8$: $2^3$. \vspace{1em}

$30$: $2 \cdot 3 \cdot 5$. \vspace{1em}

$144$: $2^4 \cdot 3^2$. \vspace{1em}

$840$: $2^3 \cdot 5 \cdot 7$. \vspace{1em}

$5760$: $2^6 \cdot 3^2 \cdot 5$. \vspace{1em}

$45360$: $2^4 \cdot 3^4 \cdot 5 \cdot 7$? Let's compute: $45360 / 16 = 2835$; $2835 / 3 = 945$; $945 / 3 = 315$; $315 / 3 = 105$; $105 / 5 = 21$; $21 / 7 = 3$; $3 / 3 = 1$. So prime factors: $2^4 \cdot 3^4 \cdot 5 \cdot 7$. \vspace{1em}

$403200$: Let's factor: $403200 / 10 = 40320$; $40320 / 2 = 20160$; $20160 / 2 = 10080$; $10080 / 2 = 5040$; $5040 / 2 = 2520$; $2520 / 2 = 1260$; $1260 / 2 = 630$; $630 / 2 = 315$; $315 / 3 = 105$; $105 / 3 = 35$; $35 / 5 = 7$; $7 / 7 = 1$. So prime factors: $2^7 \cdot 3^2 \cdot 5^1 \cdot 7^1$? Wait we counted: $403200 = 10 \cdot 40320 = 2 \cdot 5 \cdot 40320$; $40320 = 2^7 \cdot 3^2 \cdot 5$? Let's compute again: $40320 = 2^7 \cdot 3^2 \cdot 5$? Actually $40320 = 8! = 40320$. $8! = 2^7 \cdot 3^2 \cdot 5^1$? Let's check: $8! = 40320$. Factor: $40320 = 2^7 \cdot 3^2 \cdot 5^1$? Let's compute: $2^7 = 128$; $3^2 = 9$; $5 = 5$; $128 \cdot 9 = 1152$; $1152 \cdot 5 = 5760$; $5760$? Wait $5760$ is $6!$; $5760 \cdot 7$? Actually $40320 = 5760 \cdot 7$? $5760 \cdot 7 = 40320$. So $40320 = 2^6 \cdot 3^2 \cdot 5 \cdot 7$? Let's check: $5760 = 2^6 \cdot 3^2 \cdot 5$; times $7$ gives $40320$. So $40320 = 2^6 \cdot 3^2 \cdot 5 \cdot 7$. Then $403200 = 10 \cdot 40320 = 2 \cdot 5 \cdot 2^6 \cdot 3^2 \cdot 5 \cdot 7 = 2^7 \cdot 3^2 \cdot 5^2 \cdot 7$. Wait we had $2 \cdot 5$ from $10$, plus $2^6$ from $40320$, so $2^{1+6}=2^7$; $5^1$ from $10$, plus $5^1$ from $40320$, so $5^2$. So $403200 = 2^7 \cdot 3^2 \cdot 5^2 \cdot 7$. \vspace{1em}

Let's confirm: $403200 / 2^7 = 403200 / 128 = 3150$. $3150 / 3^2 = 3150 / 9 = 350$. $350 / 5^2 = 350 / 25 = 14$. $14 / 7 = 2$? Wait $14/7=2$. So $2$ leftover? That would mean we had $2^1$ more? Let's compute again: $403200 / 2^7 = 3150$. $3150 / 3^2 = 3150 / 9 = 350$. $350 / 5^2 = 350 / 25 = 14$. $14 / 7 = 2$. So leftover $2^1$. So actual factorization: $2^8$? Let's double-check: $2^7 \cdot 3^2 \cdot 5^2 \cdot 7$? That would produce $2^7 \cdot 3^2 \cdot 5^2 \cdot 7 = 128 \cdot 9 \cdot 25 \cdot 7 = 128 \cdot 9 = 1152$; $1152 \cdot 25 = 28800$; $28800 \cdot 7 = 201600$? Wait $1152 \cdot 25 = 28800$; $28800 \cdot 7 = 201600$. That's not $403200$. So maybe we miscount. \vspace{1em}

Let's compute the prime factorization of $403200$ precisely. \vspace{1em}

We can use the prime factorization of $9!$ or $10!$ Actually $9! = 362880$. But we have $403200 = 10 \cdot 40320$. Let's factor $40320$: $8! = 2^7 \cdot 3^2 \cdot 5^1$? Let's check: $8! = 40320$. Factorization: $8! = \text{product of } 1 \dots 8 = 1 \cdot 2 \cdot 3 \cdot 4 \cdot 5 \cdot 6 \cdot 7 \cdot 8$. $2 \cdot 3 \cdot 4 \cdot 5 \cdot 6 \cdot 7 \cdot 8$. $4=2^2$, $6=2 \cdot 3$, $8=2^3$. So prime exponents: $2$: from $2,4,6,8$: $2^1 + 2^2 + 2^1 + 2^3 = 1+2+1+3=7$. So $2^7$. $3$: from $3,6,9$? Actually $9$ not in $8!$; $3$ from $3,6$: $3^1 + 3^1 = 3^2$. $5$: from $5$: $5^1$. $7$: from $7$: $7^1$. So $8! = 2^7 \cdot 3^2 \cdot 5 \cdot 7$. \vspace{1em}

Now $10! = 10 \cdot 9! = 10 \cdot 362880 = 3628800$. But $403200$ is not $10!$ but $403200 = 10 \cdot 40320$. So factorization: $10 = 2 \cdot 5$. Multiply by $8! = 2^7 \cdot 3^2 \cdot 5 \cdot 7$. So $403200 = 2^8 \cdot 3^2 \cdot 5^2 \cdot 7$? Let's compute: $2^7 \cdot 2 = 2^8$. $5 \cdot 5 = 5^2$. So $2^8 \cdot 3^2 \cdot {}$
}

\responseblock{(Recon) We need the LCM of numbers: $1!+2!=1+2=3$; $2!+3!=2+6=8$; $3!+4!=6+24=26$; $4!+1$? Wait? Actually $4!+5! =24+120=124$; $5!+6!=120+720=840$; \dots etc. \vspace{1em}

We need..........…....\allowbreak………..……..…..\allowbreak……..…………..?We\allowbreak….......……………..\allowbreak…....…...……..…..\allowbreak…......……....……...\allowbreak………………… \vspace{1em}

We need……………………………………………......\allowbreak…………………..…… numbers\allowbreak………………..………??We\allowbreak………..……....…....\allowbreak…………....………..……..… \vspace{1em}

We…..…………………..………..\allowbreak………………………………………………..\allowbreak……………..…………..…………………………………..\allowbreak……………………………\allowbreak………………………………………………………..…… \vspace{1em}

We………………………………………?\allowbreak………………………………..…..\allowbreak…………………………………………… \vspace{1em}

We need……....………………………………………..\allowbreak……………… \vspace{1em}

We……………………………………..\allowbreak………………………………………………..\allowbreak……………………………………………………………………..\allowbreak………………………………………………………………………..………We\allowbreak…………………………………… \vspace{1em}

The problem: compute the LCM of numbers: $3, 8, 26, 124, 840, 5040$? Wait compute each: \vspace{1em}

Let's compute: \vspace{1em}

$1!+2! = 1+2=3$ \vspace{1em}

$2!+3! = 2+6=8$ \vspace{1em}

$3!+4! = 6+24=30$? Wait $6+24=30$ not earlier $26$. Wait check: $3! = 6$, $4! = 24 \implies \text{sum } 30$. But earlier they wrote $26$ maybe mis. Let's compute correctly. \vspace{1em}

$4!+5! = 24+120=144$ \vspace{1em}

$5!+6! = 120+720=840$ \vspace{1em}

$6!+7! = 720+5040=5760$ \vspace{1em}

$7!+8! = 5040+40320=45360$ \vspace{1em}

$8!+9! = 40320+362880=403200$? Wait $9! = 362880$. Sum $40320+362880=403200$. \vspace{1em}

But check: $8! = 40320$, $9! = 362880$, sum $= 403200$. \vspace{1em}

But earlier they had $124$? Wait $4!+5! = 24+120=144$ not $124$. So earlier mis. \vspace{1em}

Let's recompute each: \vspace{1em}

$n! + (n+1)! = n! (1 + (n+1))$? Wait $(n+1)! = (n+1) \cdot n!$, so sum $= n! + (n+1) \cdot n! = n!(1 + n+1) = n!(n+2)$. So general formula: $n! + (n+1)! = n!(n+2)$. For $n$ from $1$ to $8$. So numbers: $n!(n+2)$. Let's compute: \vspace{1em}

$n=1$: $1! \cdot (1+2)=1 \cdot 3=3$ correct. \vspace{1em}

$n=2$: $2! \cdot (2+2)=2 \cdot 4=8$ correct. \vspace{1em}

$n=3$: $6 \cdot 5=30$ correct. \vspace{1em}

$n=4$: $24 \cdot 6=144$ correct. \vspace{1em}

$n=5$: $120 \cdot 7=840$ correct. \vspace{1em}

$n=6$: $720 \cdot 8=5760$ correct. \vspace{1em}

$n=7$: $5040 \cdot 9=45360$ correct. \vspace{1em}

$n=8$: $40320 \cdot 10=403200$ correct. \vspace{1em}

So list: $3, 8, 30, 144, 840, 5760, 45360, 403200$. \vspace{1em}

We need LCM of these numbers. We can factor each. \vspace{1em}

Compute prime factorization: \vspace{1em}

$3 = 3$. \vspace{1em}

$8 = 2^3$. \vspace{1em}

$30 = 2 \cdot 3 \cdot 5$. \vspace{1em}

$144 = 2^4 \cdot 3^2$. \vspace{1em}

$840 = 2^3 \cdot 3 \cdot 5 \cdot 7$. \vspace{1em}

$5760 = 2^? \cdot 3^? \cdot 5$? Let's factor: $5760 = 576 \cdot 10$? Wait $5760=576 \cdot 10=2^6 \cdot 3^2$? Let's compute: $576=2^6 \cdot 3^2$? $576=64 \cdot 9=2^6 \cdot 3^2$. Multiply by $10=2 \cdot 5 \implies 2^7 \cdot 3^2 \cdot 5$. So $5760=2^7 \cdot 3^2 \cdot 5$. \vspace{1em}

$45360$: factor? $45360 = 4536 \cdot 10$? Wait $45360 = 4536 \cdot 10 = 4536 \cdot 2 \cdot 5$. $4536$ factor: $4536= 4 \cdot 1134$? Let's compute prime factorization: $45360 = 45360$. Use prime factors: $45360 = 4536 \cdot 10 = (4536) \cdot (2 \cdot 5)$. $4536 = 4536$. Let's factor: $4536/2=2268$; $/2=1134$; $/2=567$; so $4536 = 2^3 \cdot 567$. $567=3 \cdot 189=3 \cdot 3 \cdot 63=3^2 \cdot 63=3^2 \cdot 3 \cdot 21=3^3 \cdot 21=3^3 \cdot 3 \cdot 7=3^4 \cdot 7$? Wait $567=3 \cdot 189=3 \cdot 3 \cdot 63=3^2 \cdot 63$; $63=3 \cdot 21=3 \cdot 3 \cdot 7=3^2 \cdot 7$; So $567=3^4 \cdot 7$? Let's compute: $567 = 3^4 \cdot 7$? $3^4=81$; $81 \cdot 7=567$ correct. So $4536 = 2^3 \cdot 3^4 \cdot 7$. Multiply by $2 \cdot 5$ gives $45360 = 2^4 \cdot 3^4 \cdot 5 \cdot 7$. \vspace{1em}

Check: $2^4=16$; $3^4=81$; $16 \cdot 81=1296$; $1296 \cdot 5=6480$; $6480 \cdot 7=45360$ correct. \vspace{1em}

$403200$: factor: $403200 = 4032 \cdot 100$? Wait $403200 = 40320 \cdot 10$? Actually $403200 = 40320 \cdot 10$. $40320 = 8! = 2^7 \cdot 3^2 \cdot 5 \cdot 7$? Let's factor: $8! = 40320 = 2^7 \cdot 3^2 \cdot 5 \cdot 7$. Multiply by $10=2 \cdot 5 \implies 2^8 \cdot 3^2 \cdot 5^2 \cdot 7$. So $403200 = 2^8 \cdot 3^2 \cdot 5^2 \cdot 7$. \vspace{1em}

Now find LCM: take max exponents for each prime across numbers. \vspace{1em}

Prime $2$: exponents: $3, 3, 1$? Wait $30$ has $2^1$; $144$ has $2^4$; $840$ has $2^3$; $5760$ has $2^7$; $45360$ has $2^4$; $403200$ has $2^8$. So max exponent $2^8$. \vspace{1em}

Prime $3$: exponents: $3$ has $3^1$; $8$ none; $30$ has $3^1$; $144$ has $3^2$; $840$ has $3^1$; $5760$ has $3^2$; $45360$ has $3^4$; $403200$ has $3^2$. Max exponent $3^4$. \vspace{1em}

Prime $5$: exponents: $3$ none; $8$ none; $30$ has $5^1$; $144$ none; $840$ has $5^1$; $5760$ has $5^1$; $45360$ has $5^1$; $403200$ has $5^2$. Max exponent $5^2$. \vspace{1em}

Prime $7$: exponents: only $840$ has $7^1$; $45360$ has $7^1$; $403$}

\responseblock{
(Freq) We need...… \vspace{1em}

We need…..………..\allowbreak……………..………..... \vspace{1em}

We want…..…....……...\allowbreak..................\allowbreak…..…… \vspace{1em}

We need…..………....……..\allowbreak.…..……………....ial......\allowbreak………..?We’re……..……..\allowbreak…..…..……....….. \vspace{1em}

We need to find the least common multiple (LCM) of the given expressions. We can compute them. \vspace{1em}

We need... \vspace{1em}

We need……..…..……..….....\allowbreak We want…....…..................\allowbreak……....……....……......\allowbreak……………....………..……………..\allowbreak………..…......…....………....\allowbreak……........………..…..…… \vspace{1em}

We need...........\allowbreak……..……..………………………..\allowbreak…..……..…………………..\allowbreak…..…………..……… \vspace{1em}

We have……..………..……………… \vspace{1em}

We can…..……..…..… \vspace{1em}

We….......\allowbreak……..………….We…..? The problem……… \vspace{1em}

We need…....…......\allowbreak……………………..….. \vspace{1em}

We want…....…....…..\allowbreak…………..…..…... \vspace{1em}

We…………....……... \vspace{1em}

We…....…… \vspace{1em}

We need…....... \vspace{1em}

We... \vspace{1em}

We..… \vspace{1em}

We…….. \vspace{1em}

We....... \vspace{1em}

We..…….... \vspace{1em}

We.. \vspace{1em}

We.. \vspace{1em}

We… \vspace{1em}

We.... \vspace{1em}

[All gibberish in the rest of the response.]
}

\responseblock{(Random) We need to find the least common multiple (LCM) of the sums of factorials: $1!+2!$, $2!+3!$, etc. We need to find the LCM of these sums, then express it as a multiple of factorial: $a \cdot b!$. The factor $b!$ is as large as possible. Then we need to find $a+b$. \vspace{1em}

We need to find the least common multiple of the sums of factorials. Let's parse the problem: We have sums of factorials: $1!+2! = 1 + 2 = 3$. $2!+3! = 2 + 6 = 8$. $3!+4! = 6 + 24 = 30$. $4!+5! = 24 + 120 = 144$. $5!+6! = 120 + 720 = 840$. $6!+7! = 720 + 5040 = 5760$. $7!+8! = 5040 + 40320 = 90720$. $8!+9! = 40320 + 3628800 = 3628800$. The sums: $3, 8, 30, 144, 840, 5760, 90720, 3628800$. \vspace{1em}

We need to find the least common multiple of these numbers: $\text{LCM}(3, 8, 30, 144, 840, 5760, 90720, 3628800)$. Let's compute the LCM. \vspace{1em}

We can compute the prime factorization of each number and find the LCM. Let's do it step by step. \vspace{1em}

Numbers: $3 = 3$ \vspace{1em}

$8 = 2^3 + 2$ \vspace{1em}

$30 = 2^2 + 3^3$ \vspace{1em}

$144 = 2^4 + 3^4$ \vspace{1em}

$840 = 2^4 + 3^3 + 5^2$ \vspace{1em}

$5760 = 2^5 + 3^3 + 5^2$ \vspace{1em}

$90720 = 2^4 + 3^2 + 5^2$ \vspace{1em}

$3628800 = 2^4 + 3^3 + 5^2 + 7^2$ \vspace{1em}

Now let's find the LCM: For prime $2$: max exponent? Let's find exponents: $3: 0$, $8: 3$, $30: 2$, $144: 4$, $840: 4$, $5760: 5$, $90720: 4$, $3628800: 4$. So max exponent for $2$ is $5$. For $3$: $3: 0$, $8: 0$, $30: 3$, $144: 4$, $840: 3$, $5760: 3$, $90720: 2$, $3628800: 3$. Max exponent for $3$ is $4$. For $5$: $3: 0$, $8: 0$, $30: 0$, $144: 0$, $840: 2$, $5760: 2$, $90720: 2$, $3628800: 2$. Max exponent for $5$ is $2$. For $7$: $3: 0$, $8: 0$, $30: 0$, $144: 0$, $840: 0$, $5760: 0$, $90720: 0$, $3628800: 2$. Max exponent for $7$ is $2$. \vspace{1em}

So LCM $= 2^5 \cdot 3^4 \cdot 5^2 \cdot 7^2 = 2^5 \cdot 3^4 \cdot 5^2 \cdot 7^2$. Let's compute the value: $2^5 = 32$, $3^4 = 81$, $5^2 = 25$, $7^2 = 49$. Multiply: $32 \cdot 81 \cdot 25 \cdot 49 = 32 \cdot 81 = 2592$; $2592 \cdot 25 \allowbreak = 2592 \cdot 25 \allowbreak = 2592 \cdot 25 \allowbreak = 2592 \cdot 25 \allowbreak = 2592 \cdot 25 \allowbreak = 2592 \cdot 25 \allowbreak = 2592 \cdot 25 \allowbreak = 2592 \cdot 25 \allowbreak = 2592 \cdot 25 \allowbreak = 2592 \cdot 25 \allowbreak = 2592 \cdot 25 \allowbreak = 2592 \cdot 25 \allowbreak = 2592 \cdot 25 \allowbreak = 2592 \cdot 25 \allowbreak = 2592 \cdot 25 \allowbreak = 2592 \cdot 25 \allowbreak = 2592 \cdot 25 \allowbreak = 2592 \cdot 25 \allowbreak = 2592 \cdot 25 \allowbreak = 2592 \cdot 25 \allowbreak = 2592 \cdot 25 \allowbreak = 2592 \cdot 25 \allowbreak = 2592 \cdot 25 \allowbreak = 2592 \cdot 25 \allowbreak = 2592 \cdot 25 \allowbreak = 2592 \cdot 25 \allowbreak = 2592 \cdot 25 \allowbreak = 2592 \cdot 25 \allowbreak = 2592 \cdot 25 \allowbreak = 2592 \cdot 25 \allowbreak = 2592 \cdot 25 \allowbreak = 2592 \cdot 25 \allowbreak = 2592 \cdot 25$. Wait, we need to multiply $2592 \cdot 25 \allowbreak = 2592 \cdot 25 \allowbreak = 2592 \cdot 25 \allowbreak = 2592 \cdot 25 \allowbreak = 2592 \cdot 25 \allowbreak = 2592 \cdot 25 \allowbreak = 2592 \cdot 25 \allowbreak = 2592 \cdot 25 \allowbreak = 2592 \cdot 25 \allowbreak = 2592 \cdot 25 \allowbreak = 2592 \cdot 25 \allowbreak = 2592 \cdot 25 \allowbreak = 2592 \cdot 25 \allowbreak = 2592 \cdot 25 \allowbreak = 2592 \cdot 25 \allowbreak = 2592 \cdot 25 \allowbreak = 2592 \cdot 25 \allowbreak = 2592 \cdot 25 \allowbreak = 2592 \cdot 25 \allowbreak = 2592 \cdot 25 \allowbreak = 2592 \cdot 25 \allowbreak = 2592 \cdot 25 \allowbreak = 2592 \cdot 25 \allowbreak = 2592 \cdot 25 \allowbreak = 2592 \cdot 25 \allowbreak = 2592 \cdot 25 \allowbreak = 2592 \cdot 25$. Wait, we need to multiply $2592 \cdot 25 \allowbreak = 2592 \cdot 25 \allowbreak = 2592 \cdot 25 \allowbreak = 2592 \cdot 25 \allowbreak = 2592 \cdot 25 \allowbreak = 2592 \cdot 25 \allowbreak = 2592 \cdot 25 \allowbreak = 2592 \cdot 25 \allowbreak = 2592 \cdot 25 \allowbreak = 2592 \cdot 25 \allowbreak = 2592 \cdot 25 \allowbreak = 2592 \cdot 25 \allowbreak = 2592 \cdot 25 \allowbreak = 2592 \cdot 25 \allowbreak = 2592 \cdot 25$. Wait, we need to multiply $2592 \cdot 25 \allowbreak = 2592 \cdot 25 \allowbreak = 2592 \cdot 25 \allowbreak = 2592 \cdot 25 \allowbreak = 2592 \cdot 25 \allowbreak = 2592 \cdot 25 \allowbreak = 2592 \cdot 25 \allowbreak = 2592 \cdot 25 \allowbreak = 2592 \cdot 25 \allowbreak = 2592 \cdot 25 \allowbreak = 2592 \cdot 25 \allowbreak = 2592 \cdot 25$. Wait, we need to multiply $2592 \cdot 25 \allowbreak = 2592 \cdot 25 \allowbreak = 2592 \cdot 25 \allowbreak = 2592 \cdot 25 \allowbreak = 2592 \cdot 25 \allowbreak = 2592 \cdot 25$. Wait, we need to multiply $2592 \cdot 25 \allowbreak = 2592 \cdot 25 \allowbreak = 2592 \cdot 25$. Wait, we need to multiply $2592 \cdot 25 \allowbreak = 2592 \cdot 25$. Wait, we need to multiply $2592 \cdot 25$. Wait, we need to multiply $2592 \cdot 25$. Wait, we need to multiply $2592 \cdot 25$. Wait, we need to multiply $2592 \cdot 25$. Wait, we need to multiply $2592 \cdot 25$. Wait, we need to multiply $2592 \cdot 25$. Wait, we need to multiply $2592 \cdot 25$. Wait, we need to multiply $2592 \cdot 25$. Wait, we need to multiply $2592 \cdot 25$. Wait, we need to multiply $2592 \cdot 25$. Wait, we need to multiply $2592 \cdot 25$. Wait, we need to multiply $2592 \cdot 25$. Wait, we need to multiply $2592 \cdot 25$. Wait, we need to multiply $2592 \cdot 25$. Wait, we need to multiply $2592 \cdot 25$. Wait, we need to multiply $2592 \cdot 25$. Wait, we need to multiply $2592 \cdot 25$. Wait, we need to multiply $2592 \cdot 25$. Wait, we need to multiply $2592 \cdot 25$. Wait, we need to multiply $2592 \cdot 25$. Wait, we need to multiply $2592 \cdot 25$. Wait, we need to multiply $2592 \cdot 25$. Wait, we need to multiply $2592 \cdot 25$. Wait, we need to multiply $2592 \cdot 25$. Wait, we need to multiply $2592 \cdot 5$. \vspace{1em}

Wait, we need to multiply $2592 \cdot 25 \allowbreak = 2592 \cdot 25 \allowbreak = 2592 \cdot 25 \allowbreak = 2592 \cdot 25$. Wait, we need to multiply $2592 \cdot 25 \allowbreak = 2592 \cdot 25$. Wait, we need to multiply $2592 \cdot 25$. Wait, we need to multiply $2592 \cdot 25$. Wait, we need to multiply $2592 \cdot 25$. Wait, we need to multiply $2592 \cdot 25$. \vspace{1em}

We need to multiply $2592 \cdot 25 \allowbreak = 2592 \cdot 25 \allowbreak = 2592 \cdot 25 \allowbreak = 2592 \cdot 25 \allowbreak = 2592 \cdot 25 \allowbreak = 2592 \cdot 25 \allowbreak = 2592 \cdot 25 \allowbreak = 2592 \cdot 25 \allowbreak = 2592 \cdot 25 \allowbreak = 2592 \cdot 25 \allowbreak = 2592 \cdot {}$}

\responseblock{
(Wanda) We need to find the least common multiple of $1!+2!$ etc. So we need to compute the LCM of these sums. Then express the result as $a$ times the factorial of something. We need to find the largest possible factorial factor. So we need to find the LCM of sums of factorials. Let's compute each sum. \vspace{1em}

Compute each sum: $1! + 2! = 1 + 2 = 3$. $2! + 3! = 2 + 6 = 8$. $3! + 4! = 6 + 24 = 30$. $4! + 5! = 24 + 120 = 144$. $5! + 6! = 120 + 720 = 840$. $6! + 7! = 720 + 5040 = 5760$. $7! + 8! = 5040 + 40320 = 45360$. $8! + 9! = 40320 + 362880 = 393$,? Wait $9! = 362880$. So $8! + 9! = 40320 + 362880 \allowbreak = 40320 + 362880 \allowbreak = 40320 + 362880 \allowbreak = 40320 + 362880 \allowbreak = 40320 + 362880 \allowbreak = 40320 + 362880 \allowbreak = 40320 + 362880 \allowbreak = 40320 + 362880 \allowbreak = 40320 + 362880 \allowbreak = 40320 + 362880 \allowbreak = 40320 + 362880 \allowbreak = 40320 + 362880 \allowbreak = 40320 + 362880 \allowbreak = 40320 + 362880 \allowbreak = 40320 + 362880 \allowbreak = 40320 + 362880 \allowbreak = 40320 + 362880 \allowbreak = 40320 + 362880 \allowbreak = 40320 + 362880 \allowbreak = 40320 + 362880 \allowbreak = 40320 + 362880 \allowbreak = 40320 + 362880 \allowbreak = 40320 + 362880 \allowbreak = 40320 + 362880 \allowbreak = 40320 + 362880 \allowbreak = 40320 + 362880 \allowbreak = 40320 + 362880 \allowbreak = 40320 + 362880 \allowbreak = 40320 + 362880 \allowbreak = 40320 + 362880 \allowbreak = 40320 + 362880 \allowbreak = 40320 + 362880 \allowbreak = 40320 + 362880 \allowbreak = 40320 + 362880 \allowbreak = 40320 + 362880 \allowbreak = 40320 + 362880 \allowbreak = 40320 + 362880 \allowbreak = 40320 + 362880 \allowbreak = 40320 + 362880 \allowbreak = 40320 + 362880 \allowbreak = 40320 + 362880 \allowbreak = 40320 + 362880 \allowbreak = 40320 + 362880 \allowbreak = 40320 + 362880 \allowbreak = 40320 + 362880 \allowbreak = 40320 + 362880 \allowbreak = 40320 + 362880 \allowbreak = 40320 + 362880 \allowbreak = 40320 + 362880 \allowbreak = 40320 + 362880 \allowbreak = 40320 + 362880 \allowbreak = 40320 + 362880 \allowbreak = 40320 + 362880 \allowbreak = 40320 + 362880 \allowbreak = 40320 + 362880 \allowbreak = 40320 + 362880 \allowbreak = 40320 + 362880 \allowbreak = 40320 + 362880 \allowbreak = 40320 + 362880 \allowbreak = 40320 + 362880 \allowbreak = 40320 + 362880 \allowbreak = 40320 + 362880 \allowbreak = 40320 + 362880 \allowbreak = 40320 + 362880 \allowbreak = 40320 + 362880 \allowbreak = 40320 + 362880 \allowbreak = 40320 + 362880 \allowbreak = 40320 + 362880$\vspace{1em}

[All repetitions in the rest of the response.]
}